\documentclass[letterpaper, 10 pt, conference]{ieeeconf}  

\IEEEoverridecommandlockouts                              

\title{\LARGE \bf
A Variational Time Series Feature Extractor for Action Prediction* 
}

\author{Maxime Chaveroche$^{1}$, Adrien Malais\'e$^{2}$, Francis Colas$^{2}$, Fran\c cois Charpillet$^{2}$, Serena Ivaldi$^{2}$ 
\thanks{*This work was funded by the European Project An.Dy (no. 731540 H2020-ICT-2016-1) in the context of Maxime Chaveroche's Master thesis at INRIA in 2017.}
\thanks{$^{1}$ 
	Universit\'e de Technologie de Compi\`egne (UTC), Heudiasyc (UMR CNRS 7253)
		{\tt\small name.surname@hds.utc.fr}}%
\thanks{$^{2}$ Inria Nancy - Grand Est, Team LARSEN
        {\tt\small name.surname@inria.fr}}%
}

\usepackage{comment}
\usepackage{physics}
\usepackage{amsmath}
\usepackage{amssymb}
\usepackage{stmaryrd}
\usepackage{graphicx}
\usepackage{xcolor}
\usepackage{bm}
\usepackage{tikz-cd}
\usepackage{tikz}
\usetikzlibrary{positioning}
\usetikzlibrary{arrows}
\usetikzlibrary{shapes.geometric}
\usetikzlibrary{shapes.misc}
\tikzset{cross/.style={cross out, draw=red, minimum size=40*(#1-\pgflinewidth), inner sep=0pt, outer sep=0pt},
	cross/.default={1pt}}
\usepackage[acronym,toc]{glossaries}

\makeglossaries

\newglossaryentry{pi}
{
        name=principal investigator,
        description={Holder of an independent grant administered by a university and the lead researcher for the grant project, usually in the sciences, such as a laboratory study or a clinical trial. The phrase is also often used as a synonym for "head of the laboratory" or "research group leader"}
}

\newglossaryentry{online}
{
        name=online,
        description={Designates an algorithm able to provide results in real time, as it receives a stream of data}
}

\newglossaryentry{offline}
{
        name=offline,
        description={Designates an algorithm only able to provide results once it has all the data}
}

\newglossaryentry{one-hot}
{
        name=one-hot representation,
        description={In a multi-class classification problem, designates a class with a vector of 0 except for one component which is set to 1 and corresponds to that class}
}

\newacronym{vrnn}{VRNN}{Variational Recurrent Neural Network}

\newacronym{dnn}{DNN}{Deep Neural Network}

\newacronym{cnn}{CNN}{Convolutional Neural Network}

\newacronym{rnn}{RNN}{Recurrent Neural Network}

\newacronym{lstm}{LSTM}{Long Short-Term Memory}

\newacronym{vae}{VAE}{Variational Auto-Encoder}

\newacronym{dmp}{DMP}{Dynamic Movement Primitives}

\newacronym{dvbf}{DVBF}{Deep Variational Bayes Filters}

\newacronym{mlp}{MLP}{Multi-Layer Perceptron}

\newacronym{vi}{VI}{Variational Inference}

\newacronym{vae-dmp}{VAE-DMP}{Dynamic Movement Primitives in latent space of time-dependent Variational Auto-Encoders}

\newacronym{vtsfe}{VTSFE}{Variational Time Series Feature Extractor}

\newacronym{vbrrecopred}{VBR-Reco/Pred}{Variational Based Recurrent joint Recognition and Prediction}

\newglossaryentry{svm}
{
        name=SVM,
        description={A Support Vector Machine (SVM) is a discriminative classifier formally defined by a separating hyperplane. In other words, given labeled training data (supervised learning), the algorithm outputs an optimal hyperplane which categorizes new examples}
}

\newglossaryentry{imu}
{
        name=IMU,
        description={An Inertial Measurement Unit (IMU) is an electronic device that measures and reports a body's specific force, angular rate, and sometimes the magnetic field surrounding the body, using a combination of accelerometers and gyroscopes, sometimes also magnetometers}
}

\newglossaryentry{emg}
{
        name=Surface EMG,
        description={Surface EMG (Electromyography) assesses muscle function by recording muscle activity from the surface above the muscle on the skin. EMG recordings display the potential difference (voltage difference) between two separate electrodes. Limitations of this approach are the fact that surface electrode recordings are restricted to superficial muscles, are influenced by the depth of the subcutaneous tissue at the site of the recording which can be highly variable depending of the weight of a patient, and cannot reliably discriminate between the discharges of adjacent muscles}
}

\usepackage{url,hyperref}

\newcommand{\toremove}[1]{\textcolor{blue}{-text removed-}}

\newcommand{\removed}[1]{}
\newcommand{\colorred}[1]{{\color{red}#1\normalcolor}}
\newcommand{\equref}[1]{Eq.~\ref{#1}}

\begin{document}
\maketitle


\begin{abstract}

We propose a Variational Time Series Feature Extractor (VTSFE), inspired by the VAE-DMP model of Chen et al.~\cite{CheKarSma2016}, to be used for action recognition and prediction. Our method is based on variational autoencoders. 
It improves VAE-DMP in that it has a better noise inference model, a simpler transition model constraining the acceleration in the trajectories of the latent space, and a tighter lower bound for the variational inference.   
We apply the method for classification and prediction of whole-body movements on a dataset with 7 tasks and 10 demonstrations per task, recorded with a wearable motion capture suit.
The comparison with VAE and VAE-DMP suggests the better performance of our method for feature extraction.
An open-source software implementation of each method with TensorFlow is also provided. In addition, a more detailed version of this work can be found in the indicated code repository.
Although it was meant to, the VTSFE hasn't been tested for action prediction, due to a lack of time in the context of Maxime Chaveroche's Master thesis at INRIA.

\end{abstract}


\section{Introduction}

The problem of recognizing actions or activities has been widely addressed in the computer vision research community: it consists in the classification of a fully or partially observed action, typically observed through cameras or external motion capture \cite{Vrigkas2015survey}. 
In robotics, recognizing the human activity is paramount for enabling a proper interaction and providing assistance to the human: 
an assistive device or prosthetics could switch control modes depending on the current human activity (\textit{e.g.}, walking or sitting) \cite{Dollar2008,Varol2007}; a mobile robot may adapt its navigation depending on the prediction of the human motion \cite{Ferrer2014}. 
More generally, prediction is important to provide the robot with anticipation capabilities \cite{ivaldi2017anticipatory}. 
In collaborative robotics applications in manufacturing, such as in assembly lines, recognizing the current activity of the operator is necessary for ergonomics evaluations \cite{Akkaladevi2015industrial} and for the optimization of the robot actions. 

However, there are two critical issues that prevent the direct application of existing techniques in such scenarios.

The first issue is the availability of external sensing devices (cameras or motion captures) that poses constraints on the application for many tasks and application scenarios. 
Wearable sensors, such as the Xsens MVN Link \cite{xsenssuit}, are a valid alternative to motion capture and cameras: they offer the same tracking possibilities for the human kinematics and simplify the extraction of the body posture in case of occlusions (\textit{e.g.}, large objects or environment hiding body parts) and long distances not necessarily covered by cameras.

The second issue is that the information about the current activity may be not enough for the robot to take appropriate decisions (\textit{e.g.}, in case of physical interaction), and it may be necessary to predict also the entire trajectory for the action. 
For example, a collaborative robot may need to predict the goal and  trajectory of the human while carrying an object together \cite{Evrard2012} in order to generate appropriate coordinated actions \cite{maeda2016probabilistic}. 
This entails a prediction, classification and representation problem: we need to find a suitable representation of whole-body motions where the same features are used to recognize different actions and to predict the outcome of an action from early observations.

This presents two main challenges: i) extracting features of the action in time; ii) reducing the dimensionality of the problem to ease the prediction process, reduce overfitting and enable \gls{online} applications.
These challenges can be solved by the same process, provided that it builds a set of features that is smaller than the set of the input data dimensions, and that it contains both posture and dynamics features.
In this paper, we focus on generative models for data sequences, as these techniques aim at building representations rich enough to generate plausible variants of that data.

\emph{Dynamic Bayesian Networks} (DBN), such as \emph{Hidden Markov Models} (HMM) and Kalman filters are often a simple and effective approach. For example, in \cite{dubois2013human} they were used for fall detection, to discriminate daily activities, such as walking and sitting, from falling or lying on the ground.
HMM were also used in \cite{Mandery2016}, to automatically identify meaningful low-dimensional feature vectors for motion recognition. 
However, these model-based approaches often rely on assumptions facilitating the inference but restricting their generality.
We will thus focus on model-free approaches.

Some recent papers addressed the problem using \emph{Deep Neural Networks} (\acrshort{dnn}) trained on motion capture data.
\cite{erd} used an encoder-recurrent-decoder model to predict motions. The encoder-decoder architecture learns to reconstruct postures, while the recurrent middle layer holds the temporal dynamics. 
In \cite{CheKarSma2016}, Chen \textit{et al.} proposed a generative model for human motion generation using a deep neural architecture with \emph{\acrlong{vi}} (\acrshort{vi}) \cite{vi} and Bayesian filtering with \emph{\acrlong{dmp}} (\acrshort{dmp}) \cite{schaal2006dynamic} which ensures local space-time continuity in movement representation in a reduced space. This latent space provides plausibility in data reconstruction, while being generalizable to new samples of movements.

In \cite{butepage2017deep}, the authors compared three different generative structures of encoder-decoder networks with temporal encoding, enabling action prediction in the reduced feature space. Two used a fully connected \acrshort{dnn}, while the last one used a \emph{Convolutional Neural Networks} (\acrshort{cnn}). In all these encoder-decoder networks, the encoder learns a smaller representation of an input subsequence $x_{t:t+S}$ while the decoder learns to predict the next data subsequence $x_{t+S+1:t+2S+1}$. These models slightly outperform the results in  \cite{erd}, have lower computational complexity once trained, and are therefore applicable to online tasks, but may overfit training data due to their deterministic mapping between subsequences.

\cite{simplernn} proposed a method for motion prediction that outperforms \cite{erd} by far, and is similar to \cite{butepage2017deep}, with the exception that a noise was applied to training samples, by feeding the network with its own generated predicted sequences. This noise injection at training time prevents the system overfitting. Nevertheless, the learned representation remains biased by the application, \emph{i.e.} prediction, and thus might not learn useful features for recognition purposes. The same phenomenon may appear in \cite{vrnn}, where a \emph{\acrlong{rnn}} (\acrshort{rnn}) was employed in a generative model, alongside \emph{\acrlong{vae}}s (\acrshort{vae}) \cite{vae}, which generalizes features encoding while being biased by the integration of the RNN internal state variable.

To provide a more generalizable feature extraction, we propose a new generative model, called \emph{\acrlong{vtsfe}} (\acrshort{vtsfe}), which is task-independent since it is based on autoencoders.
Our model is inspired by VAE-DMP \cite{CheKarSma2016} and further adds three improvements:  
i) a better noise inference model, described in Section \ref{sec:noise_model}; %
ii) a simpler transition model based on a continuity constraint on the acceleration in the latent space, in contrast to the one in \cite{CheKarSma2016} represented by the DMP;
iii) a tighter lower bound (\textit{i.e.}, closer to $\ln(p_\theta(x_{1:T}))$) for the variational inference, as described in Section \ref{sec:new_lb}.
We discuss limits and benefits of the proposed method and show the results on prediction of whole-body movements 
using our open-source dataset of actions recorded with a wearable IMUs suit (Xsens MVN).

The paper is organized as follows.
Section \ref{sec:methods} introduces the related works at the base of VTSFE, described in detail in Section \ref{sec:vtsfe}.
Sections \ref{sec:materials} and \ref{sec:experiments} describe the materials and the experiments with the wearable sensors.
In Section \ref{sec:result} and \ref{conclusion} we discuss our results, conclusions and outline future works.


\section{Related works and methods}
\label{sec:methods}

Our proposed method is inspired by the work of Chen et al.~\cite{CheKarSma2016}, which proposed a method called \emph{Dynamic Movement Primitives in Latent Space of Time-Dependent Variational Autoencoders} (VAE-DMP). 
VAE-DMP is based on \emph{Deep Variational Bayes Filters} (DVBF) \cite{dvbf} that integrate Bayesian filtering into a Recurrent Deep Neural Network made of chained \emph{Variational Auto-Encoders} (VAE) \cite{vae}. VAEs  build a feature space designed to generalize postures through a reduced set of latent variables,
while Bayesian filtering ensures that time is taken into account into the spatial compression. In \cite{CheKarSma2016}, that filtering appears in the form of a \emph{Dynamic Movement Primitives} (DMP), a trajectory/dynamics model usually employed in robotics for synthesizing movement primitives \cite{schaal2006dynamic}. 

In the following, we will briefly overview the three key elements of \cite{CheKarSma2016}, namely DMP, VAE, DVBF, that are also the key methods at the base of our proposed method VTSFE . 

\subsection{DMP}

Dynamic Movement Primitives are a classical technique for learning movement primitives from demonstrated trajectories \cite{schaal2006dynamic}.
A DMP is a point attractor system written as a second-order dynamic model, aiming at representing a trajectory:
$
    \tau\ddot{p}(t) = \alpha(\beta(p_{goal} - p(t)) - \dot{p}(t)) + f
$,
where $p$ is the position, function of time, $p_{goal}$ is the goal of the trajectory, and $f$ represents the forcing term that captures the movement dynamics, 
$\tau$ is a time scaling term,
$\alpha$ and $\beta$ are constant values to tune.

\subsection{VAE}

\removed{
\begin{figure}
	\centering
	\begin{tikzpicture}[scale=0.8, every node/.style={transform shape},
		stoc/.style={draw, circle,minimum size=0.8cm, inner sep=0pt},
		det/.style={draw, diamond,minimum size=1.1cm, inner sep=0pt}
		]
	\node[draw, stoc] (x) at (-1,0) {$x$};
	\node[draw, stoc] (z) at (1,0) {$z$};
	\node[draw, stoc] (th) at (0,1) {$\theta$};
	\node[draw, stoc] (ph) at (2.5,0) {$\phi$};
	
	\draw[->,>=latex,dashed] (x) to [out=-45,in=-135] (z);
	\draw[->,>=latex,dashed] (ph) to (z);
	\draw[->,>=latex] (th) to (z);
	\draw[->,>=latex] (th) to (x);
	\draw[->,>=latex] (z) to (x);
	\end{tikzpicture}
	\caption{VAE representation consisting of the superposition of two Bayesian models. Solid lines denote the generative model with parameters $\theta$, dashed lines denote the inference model with parameters $\phi$.}
	\label{fig:vae_gm}
\end{figure}
}

A VAE \cite{vae} is an auto-encoder based on variational inference \cite{vi} that attempts to find a reduced representation of data generalizable to variants of that data.

Let $[x_k]_{k=1}^K$ be a dataset of $K$ i.i.d. samples of some continuous observation variable $x$ of unknown distribution. We assume that the data are generated by some random process, involving a latent continuous random variable $z$ and parametric families of distributions $p_{\theta}(x|z)$ and $p_{\theta}(z)$:
$    x \sim \int_z{p_{\theta^*}(x|z) p_{\theta^*}(z)dz}$,
where $\theta^*$ is the set of parameters of the parametric distribution $p_{\theta^*}(x)$. 
$\theta$ is represented by weights and biases of a \emph{decoder} neural network.
$z$ is a latent space that can be chosen arbitrarily; in VAE, the distributions are chosen to be Gaussian for simplicity:
$p_{\theta}(z) = \mathcal{N}(0,\,I)$ and $p_{\theta}(x|z) = \mathcal{N}(\mu_{x},\,\sigma_{x}^{2}I)$.

However, we don't know the distribution of $p_{\theta}(x)$, and so the distribution of $p_{\theta}(z|x)$ as $p_{\theta}(z|x) = \frac{p_{\theta}(z,x)}{p_{\theta}(x)}$. Therefore, \cite{vae} introduced a recognition model $q_{\phi}(z|x)$ designed to approximate the intractable true posterior $p_{\theta}(z|x)$. We could then constrain $q_{\phi}(z|x)$ to the same form as $p_{\theta}(x|z)$:
\begin{equation}\label{z|x}
q_{\phi}(z|x) = \mathcal{N}(\mu_{z},\,\sigma_{z}^{2}I),
\end{equation}
where $\phi$ is the set of parameters of the parametric distribution $q_{\phi}(z|x)$. 
$\phi$ is represented by weights and biases of an \emph{encoder} neural network.
This model
can then simultaneously encode and decode $x$. Yet, we have to learn $\theta$ and $\phi$ in a way that maximizes the likelihood of $[x_k]_{k=1}^K$.

Training in VAE aims both at recovering parameters $\theta$ as close as possible to the ideal parameters $\theta^*$ yielding the proper distribution over $x$, and finding parameters $\phi$ making $q_{\phi}(z|x)$ as close as possible to the intractable $p_{\theta}(z|x)$.
The latter can be achieved by minimizing the Kullback-Leibler divergence $D_{KL}(q_{\phi}(z|x)~\|~p_{\theta}(z|x))$ between these distributions.
The standard approach of variational inference is to notice that we could decompose $\ln(p_{\theta}([x_k]_{k=1}^K))$, isolate $D_{KL}(q_{\phi}(z|x_k)~\|~p_{\theta}(z|x_k))$ and minimize it by maximizing a lower bound on $\ln(p_{\theta}(x_k))$:
\begin{equation}\label{vae_max}
\ln(p_{\theta}(x_k)) = ELBO + D_{KL}(q_{\phi}(z|x_k)~\|~p_{\theta}(z|x_k))
\end{equation}
where $ELBO$ is the Evidence Lower BOund.

As the actual likelihood on the observation $p_{\theta^*}(x_k)$ is a constant and a KL-divergence is always non-negative, minimizing that KL-divergence can be done by maximizing the $ELBO$.
In terms of training the neural network, it means using $-ELBO$ as the loss function to be minimized.
This bound can in turn be split into respectively a \emph{reconstruction} (or decoder) error and a \emph{generalization} (or encoder) error:
\begin{equation}\label{vae_elbo}
\begin{aligned}
-ELBO &= - \mathbb{E}_{q_{\phi}(z|x_k)}[\ln(p_{\theta}(x_k|z))] \\
&\quad + D_{KL}(q_{\phi}(z|x_k)~\|~p_{\theta}(z))
\end{aligned}
\end{equation}

The first term of this error requires precision in reconstruction from the network by making it maximize the average likelihood $p_{\theta}(x_k|z)$ on $z$ inferred by the recognition model $q_{\phi}(z|x_k)$.
The second term of this error requires generalization from the network and balances the first error which could result in overfitting otherwise. 
The difference between VAE and a deterministic auto-encoder comes from the influence of $\sigma_{z}$ in the generalization term as it allows for the learning of a locally symmetrical and continuous latent space.
 
\subsection{DVBF}

\removed{
\begin{figure}
	\centering
	\begin{tikzpicture}[scale=0.8, every node/.style={transform shape},
	stoc/.style={draw, circle,minimum size=1cm, inner sep=0pt},
	det/.style={draw, diamond,minimum size=1.1cm, inner sep=0pt},
	rect/.style={draw, rectangle, minimum width=2.4cm,minimum height=1.2cm}
	]
	
	\node[stoc] (xt+1) at (1,-1.5) {$x_{t+1}$};
	\node[det] (zt+1) at (1,0.3) {$z_{t+1}$};
	\node[det] (zt) at (-1,0.3) {$z_t$};
	\node[stoc] (ut) at (-1,1.5) {$u_t$};
	\node[stoc] (vt) at (-2.2,-1.5) {$v_t$};
	\node[stoc] (wt) at (-1,-1.5) {$w_t$};
	\node at (-3, -1.9) {$\beta_t$};
	\node[rect] (beta) at (-1.6, -1.5) {};
	
	\draw[->,>=latex,dashed] (ut) to [out=-135,in=135] (wt);
	\draw[->,>=latex,dashed] (xt+1) to (wt);
	\draw[->,>=latex,dashed] (zt) to (wt);
	\draw[->,>=latex] (beta) to (zt+1);
	\draw[->,>=latex] (zt) to (zt+1);
	\draw[->,>=latex] (ut) to (zt+1);
	\draw[->,>=latex] (zt+1) to (xt+1);
	\end{tikzpicture}
	\caption{DVBF representation consisting of the superposition of two Bayesian models. Dependencies to $\theta$ and $\phi$ have been omitted for the sake of clarity. Solid lines denote the generative model with parameters $\theta$, dashed lines denote the inference model with parameters $\phi$. Diamond nodes indicate a deterministic dependency on parent nodes. A rectangle designates the joint distribution of the variables inside, here $\beta_t$.}
	\label{fig:dvbf_gm}
\end{figure}
}

Let us consider a set of data sequences $[[x_{k,t}]_{t=1}^T]_{k=1}^K$. 
In the following, we consider a single sample $k$ and use only the notation $\cdot _t$, as we are mostly interested in time here.

\acrshort{dvbf} \cite{dvbf} was introduced to handle time-dependency in data sequences.
Latent variables are supposed to follow a dynamics equation expressed as:
\begin{equation}\label{eq:bayes_filt}
z_{t+1} = g(z_t, u_t, \beta_t).
\end{equation}
where $g$ is a deterministic transition function, $u_t$ is a given command at time $t$ and $\beta_t$ is a set of parameters written as $\beta_t = (w_t, v_t)$, where $v_t$ are fixed universal transition parameters, and $w_t$ is a stochastic variable representing a sample-specific process noise which can be inferred from $x_t$.
In \cite{dvbf}, the authors 
 compute a variational lower bound specific to DVBF:
\begin{equation}
\begin{aligned}
&p_{\theta}(x_{1:T} | u_{1:T}, z_1) \\
&= \int_{\beta_{1:T}} p_{\theta}(\beta_{1:T}) p_{\theta}(x_t | z_1) \displaystyle \prod_{t=2}^{T} p_{\theta}(x_t | \hat{z}_t) ~d\beta_{1:T}
\end{aligned}
\end{equation}
where $\hat{z}_t=g(z_{t-1}, u_{t-1}, \beta_{t-1})$.
Then, taking the $\log$ of $p_{\theta}(x_{1:T} | u_{1:T}, z_1)$ and using Jensen's inequality, they obtain:
\begin{equation}\label{eq:dvbf_lb}
\begin{aligned}
& \ln(p_{\theta}(x_{1:T} | u_{1:T}, z_1)) \\
\geq & ~ \mathbb{E}_{q_{\phi}(\beta_{1:T}|x_{1:T}, u_{1:T}, z_1, \hat{z}_{2:T})}\left[\ln\left(p_{\theta}(x_t | z_1) \textstyle \prod_{t=2}^{T} p_{\theta}(x_t|\hat{z}_t)\right)\right]\\
& \quad - D_{KL}(q_{\phi}(\beta_{1:T}|x_{1:T}, u_{1:T}, z_1, \hat{z}_{2:T})~\|~p_{\theta}(\beta_{1:T}))
\end{aligned}
\end{equation}

This lower bound does not link to the minimization of $D_{KL}(q_{\phi}(z|x)~\|~p_{\theta}(z|x))$, but is a way to maximize the likelihood of $p_{\theta}(x_{1:T}|u_{1:T}, z_1)$, embedding transition parameters learning in the encoding process of the VAE.

\subsection{VAE-DMP}

In \cite{CheKarSma2016}, Chen et al. used \acrshort{dmp} as a dynamics model of the latent space in \acrshort{dvbf}.
They included a system noise $\epsilon_t = \epsilon \bm{w}_{\epsilon,t}$, where $\epsilon \sim \mathcal{N}(0, \Sigma_\epsilon)$, in the \acrshort{dmp} equation and defined their choice of finite difference approximations of first and second order derivatives:
\begin{equation}\label{eq:dmp_nutan}
\begin{cases}
    \tau\ddot{z}_{t+1} = \alpha(\beta(z_{T} - z_t) - \dot{z}_t) + f_t + \epsilon_t\\
	\dot{z}_{t+1} = \ddot{z}_{t+1}dt + \dot{z}_{t}\\
	z_{t+1} = z_{t+1}dt + z_t,\\
\end{cases}
\end{equation}
where $f_t$ is a continuous weighted sum of Gaussians and is deterministically inferred by a \acrlong{mlp} (\acrshort{mlp}) detailed in \cite{CheKarSma2016} that takes $x_{1:T}$ as input. This system of equations can be reshaped into the following linear form:
\begin{equation}\label{eq:trans_mat}
\begin{pmatrix}
    z_{t+1}\\
    \dot{z}_{t+1}
\end{pmatrix}
=
\begin{pmatrix}
    1 - dt^2 \alpha \beta\frac{1}{\tau} & dt^2 \alpha\frac{1}{\tau} + dt\\
    -\alpha \beta dt\frac{1}{\tau} & 1 - \alpha dt \frac{1}{\tau}
\end{pmatrix} 
\begin{pmatrix}
    z_{t}\\
    \dot{z}_{t}
\end{pmatrix}
+
b
\end{equation}
with $b=\left(dt, 1\right)^\intercal (\alpha \beta z_{T} + f_t + \epsilon_t) dt\frac{1}{\tau}$.

Contrary to standard \acrshort{vae}, in \acrshort{vae-dmp} $z_t$ is inferred by the dynamics model, along with its first-order derivative $\dot{z}_t$, at the exception of the first frame, which has no predecessor, and the last frame which is needed in \equref{eq:trans_mat}.
Therefore, the encoding process $q_{\phi}(z_1, \dot{z}_1 |x_{1:M})$, where $1 < M \leq T$, is used for the first frame, and the standard one $q_{\phi}(z_T|x_T)$ is used for the last frame of the architecture.
The noise is assumed inferred by a Gaussian distribution: $\epsilon_t \sim q_\phi(\epsilon_t|x_{t+1}, z_{t}) =\mathcal{N}(\mu_{\epsilon,t}, \Sigma_{\epsilon,t})$.

\equref{eq:trans_mat} in \acrshort{vae-dmp} replaces \equref{eq:bayes_filt} of \acrshort{dvbf}: $z_{t}$ is now $(z_{t}, \dot{z}_{t})$, the command $u_t$ of \acrshort{dvbf} is now the joint distribution of the forcing term and the goal in latent space $(f_t, z_T)$, and the transition parameters $\beta_t$ the noise $\epsilon_t$:
\begin{equation}\label{eq:chen_trans}
z_{t+1} = g(z_t, \dot{z}_{t}, z_T, f_t, \epsilon_t).
\end{equation}
As $f_{1:T}$ are deterministically inferred from $x_{1:T}$, the variational lower bound becomes:
\begin{equation}\label{vae_dmp_lb}
    \begin{aligned}
        &\ln(p_{\theta}(x_{1:T}|f_{1:T}, z_T, z_1, \dot{z}_1))\\
        &\geq \mathbb{E}_{q_{\phi}(\epsilon_{1:T}|x_{1:T}, z_T, z_1, \dot{z}_1, \hat{z}_{2:T}, \hat{\dot{z}}_{2:T})}\left[\ln(p_{\theta}(x_{1} | z_1) p_{\theta}(x_{2:T}|\hat{z}_{2:T}))\right] \\
        &\quad - D_{KL}(q_{\phi}(\epsilon_{1:T}|x_{1:T}, z_T, z_1, \dot{z}_1, \hat{z}_{2:T}, \hat{\dot{z}}_{2:T})~\|~p_{\theta}(\epsilon_{1:T}))
    \end{aligned}
\end{equation}
where we call the lower bound $\emph{ELBO}_{\emph{VAE-DMP}}$ and $q_{\phi}(\epsilon_{1:T}|x_{1:T}, z_T, z_1, \dot{z}_1, \hat{z}_{2:T}, \hat{\dot{z}}_{2:T}) = q_{\phi}(\epsilon_{1:T}|x_{1:T}, z_1, \hat{z}_{2:T})$ given their inference model.

We note that the parameters of the prior $p_{\theta}(\epsilon_t)$ are not fixed values in \cite{CheKarSma2016}, which lets the $D_{KL}$ calculation 
unclear, since its regularization power relies on fixed constraints.

\section{Proposed method: VTSFE}\label{sec:vtsfe}

In this section we describe our Variational Time Series Feature Extractor (VTSFE) for action prediction. 
The model, inspired by VAE-DMP, is shown in Figure \ref{fig:tighter_lb_gm} and has three distinguishing features:  
\textit{i)} it has a better noise inference model, described in Section \ref{sec:noise_model}; 
\textit{ii)} the transition model, that in \cite{CheKarSma2016} is represented by the DMP, here is substituted by a simple continuity constraint on the acceleration in the latent space, as described in Section \ref{sec:trans_model}; 
\textit{iii)} it has a tighter lower bound (i.e., closer to $\ln(p_\theta(x_{1:T}))$) for the variational inference, as described in Section \ref{sec:new_lb}.\footnote{Supplementary material regarding some computations (not reported for lack of space) can be found at: \url{https://github.com/inria-larsen/activity-recognition-prediction-wearable/tree/master/VTSFE}}

\subsection{Adapted noise inference}\label{sec:noise_model} 
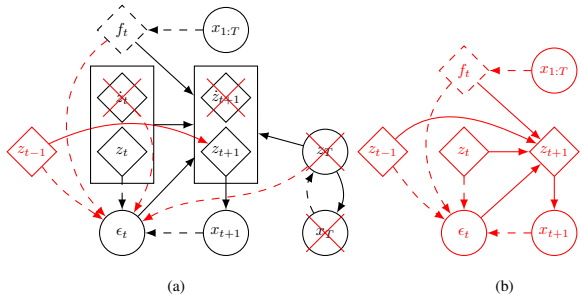
\begin{figure}
	\centering
	\begin{tikzpicture}[scale=0.6, every node/.style={transform shape},
	stoc/.style={draw, circle,minimum size=1cm, inner sep=0pt},
	det/.style={draw, diamond,minimum size=1.1cm, inner sep=0pt}
	]
	
	\node[stoc] (x1T) at (1.3,2.7) {$x_{1:T}$};
	\node[stoc] (xt+1) at (1.3,-1.8) {$x_{t+1}$};
	\node[det] (zt+1) at (1.3,0) {$z_{t+1}$};
	\node[det] (zdott+1) at (1.3,1.2) {$\dot{z}_{t+1}$};
	\node[cross] () at (1.3,1.2) {};
	\node[det] (zt) at (-1,0) {$z_t$};
	\node[det] (zdott) at (-1,1.2) {$\dot{z}_t$};
	\node[cross] () at (-1,1.2) {};
	\node[det, dashed] (ft) at (-1,2.7) {$f_t$};
	\node[stoc] (et) at (-1,-1.8) {$\epsilon_t$};
	\node[rectangle, minimum size=1.4cm](lst) at (-1,0.6) {};
	\node[rectangle, minimum size=1.4cm](lsts) at (-1,0) {};
	\node[rectangle, minimum size=1.4cm](lst+1) at (1.3,0.6) {};
	\node[stoc] (xT) at (3.5,-1.8) {$x_{T}$};
	\node[cross] () at (3.5,-1.8) {};
	\node[stoc] (zT) at (3.5,0) {$z_{T}$};
	\node[cross] () at (3.5,0) {};

	\node (a) at (0.2,-3) {(a)};
	
	\colorred{
		\node[det] (zt-1) at (-3,0) {$z_{t-1}$};
		\draw[->,>=latex] (zt-1) to [out=25,in=155] (zt+1);
		\draw[->,>=latex,dashed] (zt-1) to [out=-55,in=145] (et);
	}
	\colorred{\draw[->,>=latex,dashed] (ft) to [out=-145,in=135] (et);}
	\draw[->,>=latex,dashed] (x1T) to (ft);
	\draw[->,>=latex,dashed] (xt+1) to (et);
	\draw[->,>=latex,dashed] (zt) to (et);
	\colorred{\draw[->,>=latex,dashed] (zdott) to [out=-45, in=60] (et);}
	\draw[->,>=latex,dashed] (xT) to [out=120, in=-120] (zT);
	\colorred{\draw[->,>=latex,dashed] (zT) to [out=-140, in=30] (et);}
	\draw[->,>=latex] (zT) to (lst+1);
	\draw[->,>=latex] (zT) to [out=-60, in=60] (xT);
	\draw[->,>=latex] (lst) to (lst+1);
	\draw[->,>=latex] (et) to (lst+1);
	\draw[->,>=latex] (zt+1) to (xt+1);
	\draw[->,>=latex] (ft) to (lst+1);
	
	\draw (-1.7,-0.7) rectangle (-0.3,1.9);
	\draw (0.6,-0.7) rectangle (2,1.9);
	\end{tikzpicture}
	\begin{tikzpicture}[scale=0.6, every node/.style={transform shape},
	stoc/.style={draw, circle,minimum size=1cm, inner sep=0pt},
	det/.style={draw, diamond,minimum size=1.1cm, inner sep=0pt}
	]
	\colorred{	
		\node[stoc] (x1T) at (1.3,1.8) {$x_{1:T}$};
		\node[stoc] (xt+1) at (1.3,-1.8) {$x_{t+1}$};
		\node[det] (zt+1) at (1.3,0) {$z_{t+1}$};
		\node[det] (zt) at (-0.7,0) {$z_t$};
		\node[det] (zt-1) at (-2.5,0) {$z_{t-1}$};
		\node[det, dashed] (ft) at (-0.7,1.8) {$f_t$};
		\node[stoc] (et) at (-0.7,-1.8) {$\epsilon_t$};
	}
	
	\node (b) at (0.2,-3) {(b)};
	
	\colorred{
		\draw[->,>=latex,dashed] (ft) to [out=-135,in=125] (et);
		\draw[->,>=latex,dashed] (x1T) to (ft);
		\draw[->,>=latex,dashed] (xt+1) to (et);
		\draw[->,>=latex,dashed] (zt) to (et);
		\draw[->,>=latex,dashed] (zt-1) to [out=-55,in=145] (et);
		
		\draw[->,>=latex] (zt) to (zt+1);
		\draw[->,>=latex] (zt-1) to [out=45,in=155] (zt+1);
		\draw[->,>=latex] (et) to (zt+1);
		\draw[->,>=latex] (zt+1) to (xt+1);
		\draw[->,>=latex] (ft) to (zt+1);
	}
	\end{tikzpicture}
	\caption{Our VTSFE representation consisting of the superposition of two Bayesian models (a) compared with VAE-DMP, (b) alone. The black part is the VAE-DMP model from \cite{CheKarSma2016}, the red arrows indicate our added dependences. Red nodes and red crosses indicate respectively added variables and removed variables in our final model (b).
		Dependencies to $\theta$ and $\phi$ have been omitted for the sake of clarity. Solid lines denote the generative model with parameters $\theta$, dashed lines denote the inference model with parameters $\phi$. Diamond nodes indicate a deterministic dependency on parent nodes. Diamond dashed node indicates a deterministic dependency on parent nodes for the inference model but not the generative one. A rectangle designates the joint distribution of the variables inside, here $(z_t, \dot{z}_t)$.}
	\label{fig:tighter_lb_gm}
\end{figure}

\acrshort{vae-dmp} uses the inference model of DVBF.
However, the change of variable between $z_t$ and $(z_t, \dot{z}_t)$ and the dependency on $z_T$ of the transition model should affect the noise inference, which is supposed to fill the gap remaining between information contained in the latent space and the one in observation space. Besides, knowing $x_{t+1}$ alone isn't enough to deterministically infer $f_t$.
Therefore, we propose the new following noise inference:
$
\epsilon_t \sim q_\phi(\epsilon_t|f_t, x_{t+1}, z_{t}, \dot{z}_t, z_T),
$
if one chooses to use $\dot{z}_t$, or:
$
\epsilon_t \sim q_\phi(\epsilon_t|f_t, x_{t+1}, z_{t}, z_{t-1}, z_T),
$
otherwise.
The corresponding Bayesian models are illustrated in Figure \ref{fig:tighter_lb_gm}.

Now, let us assume that the prior $p_{\theta}(\epsilon_t)$ of our model is a Gaussian white noise that does not depend on $t$. 
We consider $\sigma_{scale}$ the scaling term for all noise $\epsilon_{t}$, that also affects its inference mean knowing $x_{t}$. We could then make the following assumptions:
\begin{equation}\label{eq:noise_assumptions}
\begin{aligned}
q_{\phi}(\epsilon_{t}|f_{t}, x_{t+1}, z_{t}, z_{t-1}, z_{T}) &=  \mathcal{N}(\sigma_{scale} \cdot \mu_{\epsilon,t}, ~ \sigma_{scale}^2 \sigma_{\epsilon,t}^2I)\\
p_{\theta} (\epsilon_t) &=  \mathcal{N}(0, \sigma_{scale}^2I).
\end{aligned}
\end{equation}
Actually, as $\mu_{\epsilon,t}$, $\sigma_{scale}$ and $\sigma_{\epsilon,t}$ are all inferred by our neural network, this formulation is equivalent to:
\begin{equation}\label{eq:noise_assumptions_no_scale}
\begin{aligned}
q_{\phi}(\epsilon_{t}|f_{t}, x_{t+1}, z_{t}, z_{t-1}, z_{T}) &=  \mathcal{N}(\mu_{\epsilon,t}, ~ \sigma_{\epsilon,t}^2I)\\
p_{\theta} (\epsilon_t) &=  \mathcal{N}(0, \sigma_{scale}^2I),
\end{aligned}
\end{equation}
but leads to a lighter Kullback-Leibler divergence expression.

\subsection{Lighter transition model}\label{sec:trans_model}

In \cite{CheKarSma2016} \acrshort{dmp}s were used as a space-time continuity constraint, i.e., a constraint that gives to the network the notion of continuous acceleration. However, 
we would like to remove the dependency to the point-attractor aspect of \acrshort{dmp}, remove the hyperparameters $\alpha$, $\beta$ and $\tau$, as well as the appearance of $z_{T}$ in its equation. Thus, in association with the central finite difference approximator of acceleration (which is  more accurate than the backward and forward variants), \equref{eq:dmp_nutan} simply becomes:
\begin{equation}
\begin{aligned}
\begin{cases}
\ddot{z}_{t} = f_t + \epsilon_t\\
\ddot{z}_{t} = \frac{z_{t+1} - 2 z_t + z_{t-1}}{dt^2}
\end{cases}
\end{aligned}
\end{equation}
\equref{eq:trans_mat} then becomes:
$z_{t+1} = (f_t + \epsilon_t) dt^2 + 2 z_t - z_{t-1}$,
where $dt$ doesn't even act as a hyperparameter neither since $f_t$ and $\epsilon_t$ are learned and completely artificial.
The new transition models, formerly \equref{eq:chen_trans}, becomes:
$z_{t+1} = g(z_t, z_{t-1}, f_t, \epsilon_t)$
and it has several benefits.
 First, it removes the model parametrization, which prevents us from doing an expensive grid search. Then, it greatly alleviates the complexity of our model both at transition time and at loss computation time, since it removes the need to sample the $z_{T}$ prior.

\subsection{Lower bound}\label{sec:new_lb}

In this subsection, we will define a new lower bound for our \acrshort{vtsfe} model to use as loss function. We will make the calculation with the full \acrshort{dmp} transition model to obtain a general lower bound, whether we use the point-attractor aspect or not.
Contrary to \cite{CheKarSma2016}, we will not use the \acrshort{dvbf} lower bound, as the main difference in \acrshort{vtsfe} and \acrshort{vae-dmp} with respect to \acrshort{dvbf} is that unknown forcing term $f_t$ replace observed commands $u_t$.
This means that we could look for a different variational lower bound using $\ln(p_{\theta}(x_{1:T}))$ rather than $\ln(p_{\theta}(x_{1:T}|f_{1:T}, z_T, z_1, \dot{z}_1)))$ as in \equref{vae_dmp_lb}.
Similarly to \equref{vae_elbo}, we can write:
\begin{equation}\label{eq:new_elbo}
    \begin{aligned}
        ELBO &= \mathbb{E}_{q_{\phi}(z_{1:T}|x_{1:T})}[\ln(p_{\theta}(x_{1:T}|z_{1:T}))]\\
        &\quad -D_{KL}(q_{\phi}(z_{1:T}|x_{1:T})~\|~p_{\theta}(z_{1:T}))
    \end{aligned}
\end{equation}

Instead of inserting the derivative in the state so as to get a first-order Markov assumption, we simply use the second-order Markov property.
With a VAE encoder only for the first, second and last time steps, we can show that:
\begin{equation}\label{eq:new_lb_reco_err}
\begin{aligned}
&\mathbb{E}_{q_{\phi}(z_{1:T}|x_{1:T})}[\ln(p_{\theta}(x_{1:T}|z_{1:T}))]\\
=&\sum\textstyle_{t\in \lbrace{1,2,T}\rbrace} \mathbb{E}_{q_{\phi}(z_{t}|x_{t})}[ \ln(p_{\theta}(x_{t}|z_{t}))] \\
&+ \sum\textstyle_{t=3}^{T-1} \mathbb{E}_{q_{\phi}(z_{1:t},z_T|x_{1:T})}[ \ln(p_{\theta}(x_{t}|z_{t}))],\\
\end{aligned}
\end{equation}
and:
\begin{equation}\label{eq:dkl_decomp}
\begin{aligned}
& D_{KL}(q_{\phi}(z_{1:T}|x_{1:T})~\|~p_{\theta}(z_{1:T}))\\
=& \sum\textstyle_{t\in \lbrace{1,2,T}\rbrace} D_{KL}(q_{\phi}(z_{t}|x_{t})~\|~p_{\theta}(z_{t})) \\
&+ \sum\textstyle_{t=2}^{T-2} \mathbb{E}_{q_{\phi}(z_{1:t}, z_T|x_{1:T})} \left[ - \ln(p_{\theta}(\hat{f}_{t})) \right.\\
& \left. \qquad + D_{KL}(q_{\phi}(\epsilon_{t}|\hat{f}_{t}, x_{t+1}, z_{t}, z_{t-1}, z_{T})~\|~p_{\theta}(\epsilon_{t}))\right],\hspace{-10pt}
\end{aligned}
\end{equation}
where $\hat{f}_t$ is the deterministic inferred value of $f_t$ for a given $x_{1:T}$.
Then, following \cite{kl} and using \equref{eq:noise_assumptions}, we get:
$D_{KL}(q_{\phi}(\epsilon_{t}|f_{t}, x_{t+1}, z_{t}, z_{t-1}, z_{T})~\|~p_{\theta}(\epsilon_{t})) $
$= \frac{1}{2} \sum_{i=1}^{d_z} \left[ - \ln(\sigma_{\epsilon,t,i}^2) - 1 + \sigma_{\epsilon,t,i}^2 + \mu_{\epsilon,t,i}^2\right]$.

Using the same discrete approximation of the mean as in \cite{vae}, 
this leads us to a computationally highly expensive approximation in $\Theta(L^{T-3})$. In order to be able to compute that new lower bound, we made a rougher approximation by taking the $L^3$ samples from the three priors $p_{\theta}(z_1)$, $p_{\theta}(z_2)$ and $p_{\theta}(z_{T})$, and for all $t \in [2, T-2]$, only propagating $z_{t+1} = g(\mu_{z,T}, \mu_{z,t}, \mu_{z,t-1}, \mu_{\epsilon,t}, \hat{f}_{t})$ for each of these $L^3$ 3-tuples $(z_{l_1},z_{l_2},z_{l_T})$, which changes \equref{eq:noise_assumptions} to:
\begin{equation}
\begin{aligned}
 &q_{\phi}(\epsilon_{t}|f_{t}, x_{t+1}, z_{t}, z_{t-1}, z_{T}) \\
=&\mathcal{N}\left(\sigma_{scale} \cdot \mu_{\epsilon,t}, ~ \sigma_{scale}^2 \left( \sigma_{\epsilon,t}^2 + K^2 \sum\textstyle_{t'=2}^{t-1} \sigma_{\epsilon,t'}^2 \right) I\right)\\
&p_{\theta}(\epsilon_t) =  \mathcal{N}(0, [1 + (t-2) K^2] \sigma_{scale}^2 I)
\end{aligned}
\end{equation}
as $g$ is a linear transformation. $K$ is the real factor that multiplies $\epsilon_{t}$ in $g$.
Thereby, we obtain that 
$D_{KL}(q_{\phi}(\epsilon_{t}|f_{t}, x_{t+1}, z_{t}, z_{t-1}, z_{T})~\|~p_{\theta}(\epsilon_{t}))$
 leads to $\sum_{t=2}^{T-2} L^3$ samples and thus a $\Theta(L^3)$ complexity to approximate the expectation of $D_{KL}$.
That approximation has also been used on \equref{eq:new_lb_reco_err} with 4-tuples $(z_{l_1}, z_{l_2}, z_{l_t}, z_{l_T})$ instead of the previous 3-tuples in order to sample the latent space at each inference on $z$ as in VAE, leading to $\Theta(L^4)$.

Also concerning \equref{eq:new_lb_reco_err}, it is worth mentioning that $\sigma_{x,t}$ is only used in the reconstruction loss term, without any other constraint. Therefore, it acts as a degree of freedom at optimization time that can compensate errors either by being high to reduce the squared error or by being very close to zero to make the log take a negative value if the error is small, which in particular could happen if the input data range doesn't vary much around zero, as it is the case for our data that ranges in $[-1, 1]$. Besides, having a different $\sigma_{x,t}$ for each $z$ sample causes the optimization process to treat each error differently which reinforces these issues.
Thus, in order to correct that error compensation and to save gradient computations, we made an additional assumption on $p_{\theta}(x_t | z_t)$ by setting $\sigma_{x,t}^2 = \frac{1}{2\pi}$, where the value $\frac{1}{2\pi}$ has been chosen to remove all constant term so the reconstruction error is entirely due to the squared term. This new assumption allows us to replace the expression of $\ln(p_{\theta}(x_t | z_t))$ 
with:
\begin{equation}\label{eq:likelihood_whithout_sigma}
\ln(p_{\theta}(x_t | z_t)) = \pi (x_{t,i} - \mu_{x,t,i})^{2}
\end{equation}

Let us now define the last term of our lower bound. Unfortunately, $p_{\theta}(\hat{f}_{t})$ is intractable. We could simply continue without it though, giving us a lower bound of $D_{KL}(q_{\phi}(z_{1:T}|x_{1:T})~\|~p_{\theta}(z_{1:T}))$. Nevertheless, considering it as an important loss term on dynamics reconstruction, we want to keep at least a part of it. We notice that, being the result of a Kullback-Leibler divergence with a Dirac distribution, it acts as a penalty if $P_{\theta}(\hat{f}_{t}) \neq 1$. However, given the fact that without any knowledge, $f_{t}$ could be equally anything, as $x_t$, $p_{\theta}(f_{t})$ should be closer to a uniform distribution over an infinite space than a Dirac distribution centered on a particular $\hat{f}$. Thus, we assume that $-\mathbb{E}_{q_{\phi}(z_{1:T}|x_{1:T})}[\ln(p_{\theta}(\hat{f}_{t}))] \geq -\mathbb{E}_{q_{\phi}(z_{1:T}|x_{1:T})}[\ln(p_{\theta}(\hat{f}_{t} | z_{1:T}))] \geq 0$ since $\hat{f}_{t}$ is used to infer $z_{1:T}$. 
So we need to define $p_{\theta}(f_{t}| z_{1:T})$. First, we make the assumption that it follows a Gaussian distribution.
Then, given the deterministic transition function $z_{t+1} = g(z_{l_{T}}, z_{t}, z_{t-1}, \epsilon_t, f_{t})$, where $\epsilon_t$ is simply added to $f_t$, we notice that the only variance contained in $p_{\theta}(f_{t}| z_{1:T})$ is the one of $p_{\theta}(\epsilon_t)$. Besides:
\begin{equation}
\begin{aligned}
p_{\theta}(f_{t} | z_{1:T}) &= \int_{x_{1:T}} p_{\theta}(f_{t} | x_{1:T}, z_{1:T}) p_{\theta}(x_{1:T} | z_{1:T}) ~dx_{1:T} \\
&\approx \int_{x_{1:T}} q_{\phi}(f_{t} | x_{1:T}) p_{\theta}(x_{1:T} | z_{1:T}) ~dx_{1:T}.
\end{aligned}
\end{equation}

Thereby, we make the assumption that $p_{\theta}(f_{t} | z_{1:T}) \approx \mathcal{N}(d(x_{1:T})_t,\, [1 + (t-2) K^2] \sigma_{scale}^2 I)$, where $x_{1:T} \sim p_{\theta}(x_{1:T} | z_{1:T})$ and $d$ represents a function taking $x_{1:T}$ as input and $f_{1:T}$ as output where $f_{1:T} \sim q_{\phi}(f_{t} | x_{1:T})$, as the MLP used to infer $f_{1:T}$. 

Unfortunately, even combined with the previously defined $\Theta(L^3)$ approximation of $\mathbb{E}_{q_{\phi}(z_{1:T}|x_{1:T})}$, the additional sampling on $x_{1:T}$ still leads to a $\Theta(L^{3}\times X^T)$ approximation, where $X$ is the number of samples taken for $x_t$. Therefore, we made an even rougher approximation by taking $X=1$, where that only sample is the T-tuple $(\mu_{x,t})_{t \in [1:T]}$ and $\mu_{x,t}$ is the mean of $p_{\theta}(x_t|z_t)$. 

Finally, we obtain the following relations on the various Evidence Lower BOunds (\emph{ELBO}):
\begin{displaymath}
    \emph{ELBO} \geq \emph{ELBO}_{\emph{VTSFE}} \geq \emph{ELBO}_{\emph{VAE-DMP}}
\end{displaymath}

More precise mean approximations aside, our lower bound $\emph{ELBO}_{\emph{VTSFE}}$ is thereby tighter to the initial \emph{ELBO} than the DVBF one from which derives $\emph{ELBO}_{\emph{VAE-DMP}}$ thanks to our \emph{dynamics reconstruction loss term} $-\mathbb{E}_{q_{\phi}(z_{1:T}|x_{1:T})}[\ln(p_{\theta}(\hat{f}_{t} | z_{1:T}))]$.

However, due to our approximations, we end up with a complexity, given $T$, of $\Theta(L^4)$ (or more precisely of $\Theta(P^3\times M)$, where $P$ is the number of samples from each of the priors $(z_{l_1}, z_{l_2}, z_{l_T})$ to propagate through time, and $M$ is the number of samples from $\epsilon_{t-1}$ at time $t$).

That new lower bound is more computationally expensive, but should allow the creation of a more generalizable latent space by creating local space continuity in terms of both surroundings of each input and trajectory of the whole sequence, i.e. reducing the intraclass variance in latent space for each movement type, for $L^3$ trajectory initialization samples (prior sampling). At the same time, 
it forces the reconstructed inputs $x_{1:T}$ to keep the same dynamics as the actual inputs, in addition to force them to be individually close to the actual ones.

To reduce the complexity, we propose a \emph{light VTSFE} by making a rough $\Theta(L)$ approximation consisting of only one prior sampling $(\mu_{z,1}, \mu_{z,2}, \mu_{z,T})$ for $D_{KL}$ 
and only $L$ samples $(\mu_{z,1}, \mu_{z,2}, \mu_{z,T}, z_{l_{t}})$ for
$-\mathbb{E}_{q_{\phi}(z_{[1:t]},z_T|x_{1:t})}[ln(p_{\theta}(x_{t}|z_{t}))]$. 
This version 
reduces intraclass variance in latent space for each movement type, but only processes one trajectory initialization sample (prior sampling).

\subsection{Neural network architecture}\label{sec:nn}

\subsubsection{Differences with respect to the original VAE}
We used a Gaussian MLP (as defined in appendix C of \cite{vae}) as encoder to model $q_{\phi}(z|x)$, and another Gaussian MLP as decoder to model $p_{\theta}(x|z)$.
Instead of \emph{tanh} as activation function for hidden layers, we used \emph{elu} (Exponential Linear Unit) \cite{elu}, as it outperforms state-of-the-art activation functions in learning speed and accuracy, and alleviates the vanishing gradient problem. 
We used the ADAM optimization algorithm \cite{adam} instead of ADAGRAD \cite{adagrad}. 
\subsubsection{Sequence architecture}

Since we want to build a unique latent space for all movement frames, all network variables are shared between time steps.
As in \cite{CheKarSma2016}, because of the vanishing gradient problem and the lack of movement demonstrations (but mostly because we want the original VAE to process all inputs and only use the DMP equation as a constraint), we split our samples into overlapping subsequences of size $l_{sub}$, where the architecture of our neural network is built using $T = l_{sub}$. The overlap is equal to $l_{sub} - n_{V}$, where $n_{V}$ is the number of standard VAEs.
However, we used the whole data sequence to infer $f_{1:T}$ given $x_{1:T}$ in order to remove a degree of freedom at optimization time between subsequences; we used \emph{elu} instead of \emph{softmax} as activation function in the MLP that infers $f_{1:T}$ for the same reasons than above.

\subsection{Loss function gradients}

Learning occurs through the back-propagation algorithm given a loss function and its gradients. 
To ensure precision, all gradient computations were made atomically, i.e. per time step and per dimension, but the back-propagation algorithm always was executed per subsequence.
There are several ways to compute these gradients: simply computing the gradients from the whole loss function, computing them from subparts of that loss function, computing gradients for only a subset of variables, etc. Here, we considered two main learning schemes for the \acrshort{vtsfe} learning.
The first, called ``\emph{all}'', simply computes gradients for all variables from the whole loss function. It was applied to the chain of independent VAEs.
The second, called ``\emph{separated}'', was applied to VTSFE and our implementation of VAE-DMP.
First, \emph{separated} computes gradients for all variables from the recognition loss term, as it takes account for the whole system leading to the reconstruction of $x_t$. 
Then, it computes gradients from the generalization loss term for only encoder variables on which it depends. i.e. not $\sigma_{scale}$
Finally, it computes gradients from the dynamics loss term for a subset of variables composed of all variables except the weights that define the forcing term $f_t$ and the noise scaling variable $\sigma_{scale}$, since the dynamics loss term is supposed to only influence the quality of the decoder and, as a side effect, the one of the $\epsilon_t$ encoder; It should not try to optimize directly the variables used to compute its loss, i.e. Gaussian basis weights defining $f_t$ and $\sigma_{scale}$, to avoid optimization shortcuts.

\section{Materials}
\label{sec:materials}

\begin{figure}[t]
\includegraphics[width=1\hsize]{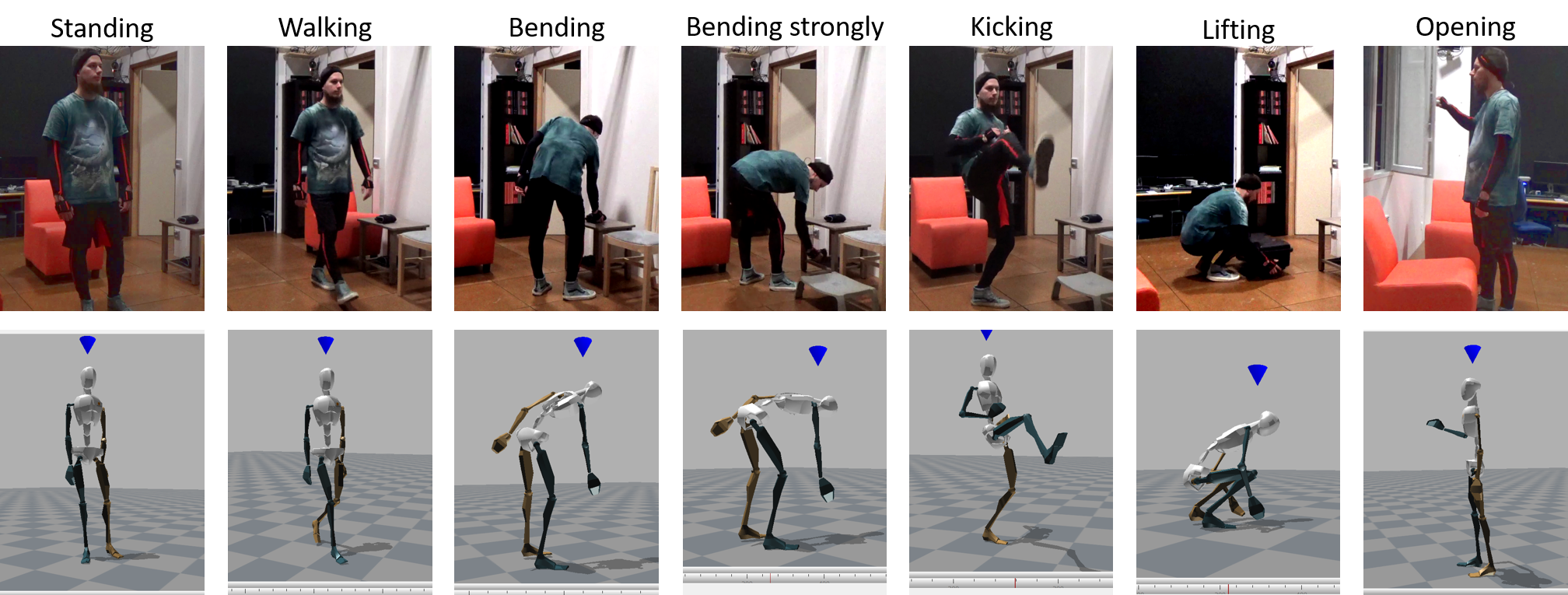}
\caption{Top: the actions of our dataset (we have 10 demonstrations per action). Bottom: the kinematics estimation of the human posture during the actions, in Xsens MVN studio.}
\label{fig:adrien}
\vspace{-0.8cm}
\end{figure}

\subsubsection{Equipment}

To capture human motion, we use an Xsens MVN inertial motion capture suit \cite{xsenssuit}, a wearable system with 17 wireless IMUs embedded in a lycra suit, tracking  motion at 240Hz. IMUs data are mapped on a calibrated Xsens biomechanical model with 23 segments, connected via 22 three-dimensional joints.
Our sequences thereby have 66 dimensions.


\subsubsection{Dataset}

To compare our VAE-DMP implementation with the original, we retrieved from the authors of \cite{CheKarSma2016} their dataset. It consisted of 5 sample trajectories, one per movement type (kicking, taïchi, balancing, walking, punching), consisting of body joint angles recorded from a motion capture system. 
To develop our model, we recorded a new dataset with our wearable sensing suit. 
Our dataset is made of 10 demonstrations per 7 movement types, of about 2 seconds.
These movements were chosen for either their similarity with each others, their complexity (the more the limbs are in movement, the more difficult it is to make compromises in a representation common to several movement) and/or their relevance in the context of industrial scenarios. These movements are: bent forward to move an object, strongly bent forward, lifting a box from the ground, kicking, opening a window, walking and standing (see Figure \ref{fig:adrien}).

\subsubsection{Software}

Our software and dataset are open-source and available on github at: \url{https://github.com/inria-larsen/activity-recognition-prediction-wearable}.

\section{Experiments} \label{sec:experiments}

We consider the following four models:
\renewcommand{\labelenumi}{\Roman{enumi}}
\makeatletter
\newcommand*{\rom}[1]{(\expandafter\@slowromancap\romannumeral #1@)}
\makeatother
\begin{enumerate}
\item Our $\Theta(L^3)$ VTSFE model, with tighter lower bound and no attractor point,
\item A $\Theta(L)$ light variant of \rom{1}, which only considers the trajectory resulting from one single sample from the $z$ priors, \textit{i.e.} $(\mu_{z,1}, \mu_{z,2})$,
\item Our $\Theta(L)$ implementation of \cite{CheKarSma2016} \footnote{We use their lower bound on the trajectory composed of $(\mu_{z}, \mu_{\dot{z}})_{1:T}$ with the DMP attractor point system and without the dynamics reconstruction loss term, our $p_{\theta}(\epsilon_t)$, our Gaussian log likelihood without variance (\equref{eq:likelihood_whithout_sigma}), their noise inference model and without their annealing schedule described in \cite{CheKarSma2016}, since our additional assumptions and implementation resolved the problem of $\epsilon_t$ acting as a shortcut to encode dynamics information.},
\item Independent standard \acrshort{vae}s trained with overlapping subsequences.
\end{enumerate}

Unfortunately, \rom{1} is computationally very expensive, to the point that we could not test it with cross-validation. Early tests did not show a significant improvement w.r.t. the lighter version \rom{2}.
Thus, we focus our comparison on \rom{2}, \rom{3} and \rom{4}.
All models were trained in the same manner to avoid training biases: samples were split into overlapping subsequences of size $l_{sub} = 10$, as in \cite{CheKarSma2016}, and an overlap size equal to $l_{sub} - n_{V}=8$, where $n_{V}$ is the number of standard VAEs. 
As in \cite{CheKarSma2016}, the encoder and decoder networks were all made of one hidden layer of 200 units. We implemented a middle layer of 10 hidden units for each frame observed as input in the MLP designed to infer $f_t$. $f_t$ was composed of 50 basis Gaussian functions, 
parametrized by standard deviation of 2.5, experimentally chosen following a grid search based on reconstruction error.

For \rom{3}, the parameters $\alpha$, $\beta$, $\tau$ and $dt$ were experimentally chosen after a grid search based on reconstruction error, with the same constraint as in \cite{CheKarSma2016}, \textit{i.e.} $\beta = \alpha/4$. We used the values: $\alpha = 2$, $\beta = 0.5$, $\tau = 1/2$ and $dt = 0.5$.

To increase learning speed and stability, we trained all models using a mini-batch of 7, again chosen following a grid search under the constraint of being a multiple of the number of training samples. 
For all models and for each VAE, we set the number of Monte Carlo samples 
at $L = 30$ to have a statistically significant sample group, as our batch size was small \cite{vae}.

For the experiments, we dynamically subsampled our dataset to fit in 70 frames of movements as in \cite{CheKarSma2016}. We also normalized the joint angles data between $[-1, 1]$.
The whole dataset is uniformly shuffled before each model training to avoid successive overfitting, and thus ensure learning convergence.

\section{Results}\label{sec:result}

\begin{table}[t]
	\centering
	\begin{tabular}{|l|c|r|}
	  \hline
	   & MSE $\times 10^{-3}$ & $\sum var$ \\
	  \hline
	  VTSFE light \rom{2} 2D & 10.195466 & $\approx$ \textbf{10} \\
	  VAE-DMP \rom{3} 2D & 9.2731696 & $\approx$ 5 \\
	  VAE only \rom{4} 2D & \textbf{8.0171376} & $\approx$ 5.5 \\
	  \hline
	  VTSFE light \rom{2} 5D & 9.8597025 & $\approx$ \textbf{9.5} \\
	  VAE-DMP \rom{3} 5D & 8.5839415 & $\approx$ 3.5 \\
	  VAE only \rom{4} 5D & \textbf{8.3727958} & $\approx$ 6 \\
	  \hline
	  VTSFE light \rom{2} 7D & 9.835138 & $\approx$ \textbf{9.5} \\
	  VAE-DMP \rom{3} 7D & 8.4773144 & $\approx$ 3.5 \\
	  VAE only \rom{4} 7D & \textbf{8.1229284} & $\approx$ 5.5 \\
	  \hline
	\end{tabular}
	\caption{Model comparison for a 2-5-7-dimensional latent space}
	\vspace{-0.8cm}
	\label{fig:results}
\end{table}

\begin{figure}[t]
  \centering
  {
  
(a)  \includegraphics[scale=0.15]{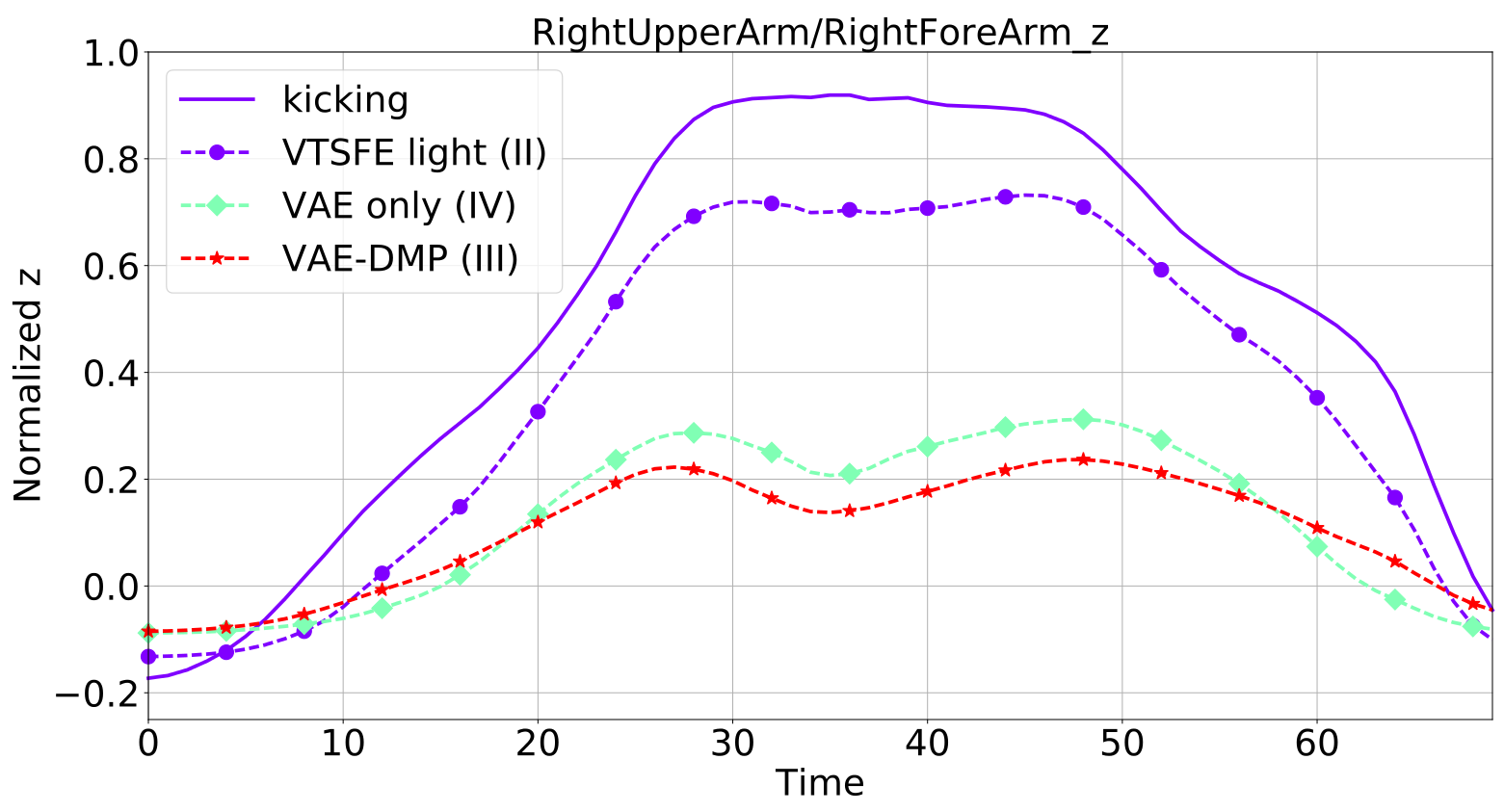}

(b)  \includegraphics[scale=0.15]{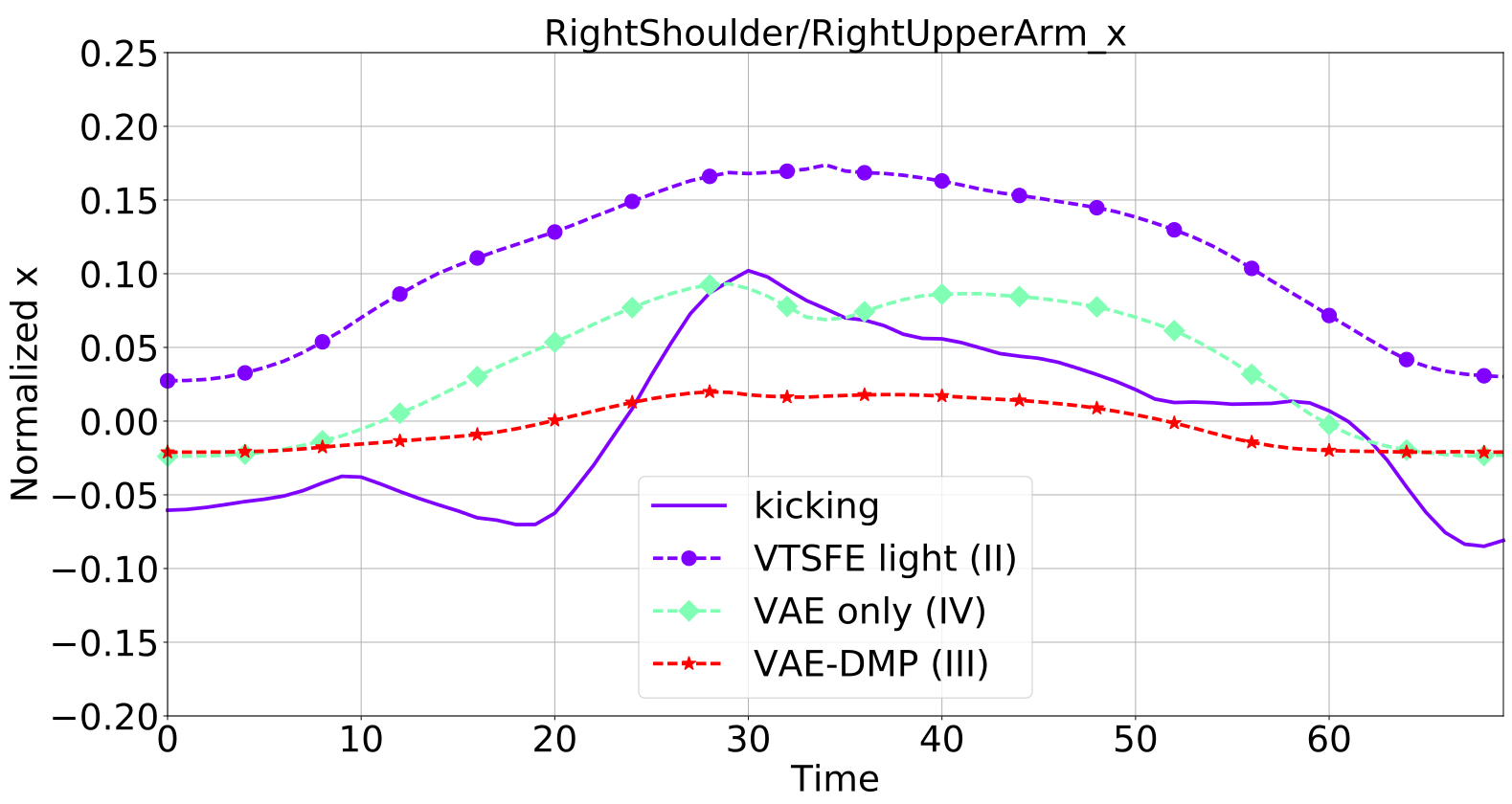}
  
  }
  \caption{Reconstruction of a new movement sample from a (a) 2D (b) 7D latent space. The solid line denotes the ground truth curve corresponding to the evolution through time of the normalized (a) $z$ spatial dimension of the right elbow (b) $x$ spatial dimension of the right shoulder angle of the human during a movement of type ``kicking''. Dashed lines denote reconstructed data. The purple one corresponds to VTSFE light \rom{2}.
  The red one corresponds to VAE-DMP \rom{3}.}
  \vspace{-0.5cm}
  \label{fig:dynamics_both}
\end{figure}

We compared the three models based on two metrics: the Mean Squared Error (MSE) between ground truth and reconstruction, and the variance of the reconstruction through time summed on all dimensions and movement samples ($\sum var$). The first one is a precision metric on the reconstruction, while the second is an indicator of the amplitude of reconstructed dynamics. Together, they provide information on the features encoded in the latent space. We averaged these values after the execution of a leave-one-out on sample indices (in the range [0,9]) which led to training on 63 samples and testing on 7 samples. These results can be found in Table \ref{fig:results}.

Our results show that VTSFE light \rom{2} leads to a slightly lower mean precision of the reconstruction error, despite being almost equivalent to VAE-DMP \rom{3} and \rom{4}, while significantly encoding more dynamics information. This increase in dynamics precision is illustrated in Figure \ref{fig:dynamics_both}a. The reason why \rom{3} and \rom{4} still reach the best precision is because of static errors of \rom{2} and movement averaging from \rom{3} and \rom{4}, as illustrated in Figure 
\ref{fig:dynamics_both}b.
Also, while precision slightly increases with the number of latent variables for all models (except VAE only), the dynamics encoding is almost constant (except VAE-DMP which decreases in quality).

\begin{figure}[ht]
  \centering
  {
(a)  \includegraphics[scale=0.16]{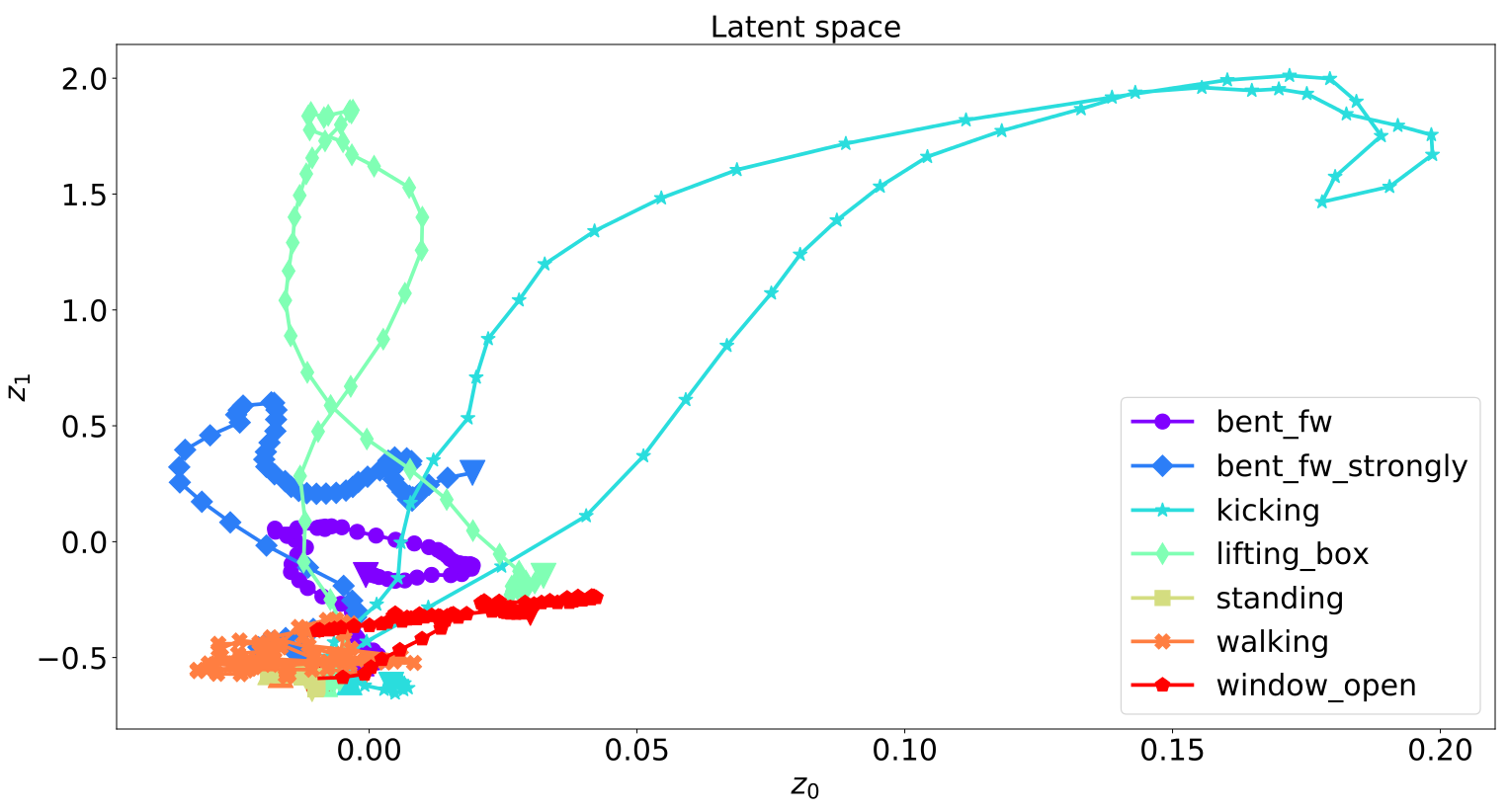}

(b)  \includegraphics[scale=0.16]{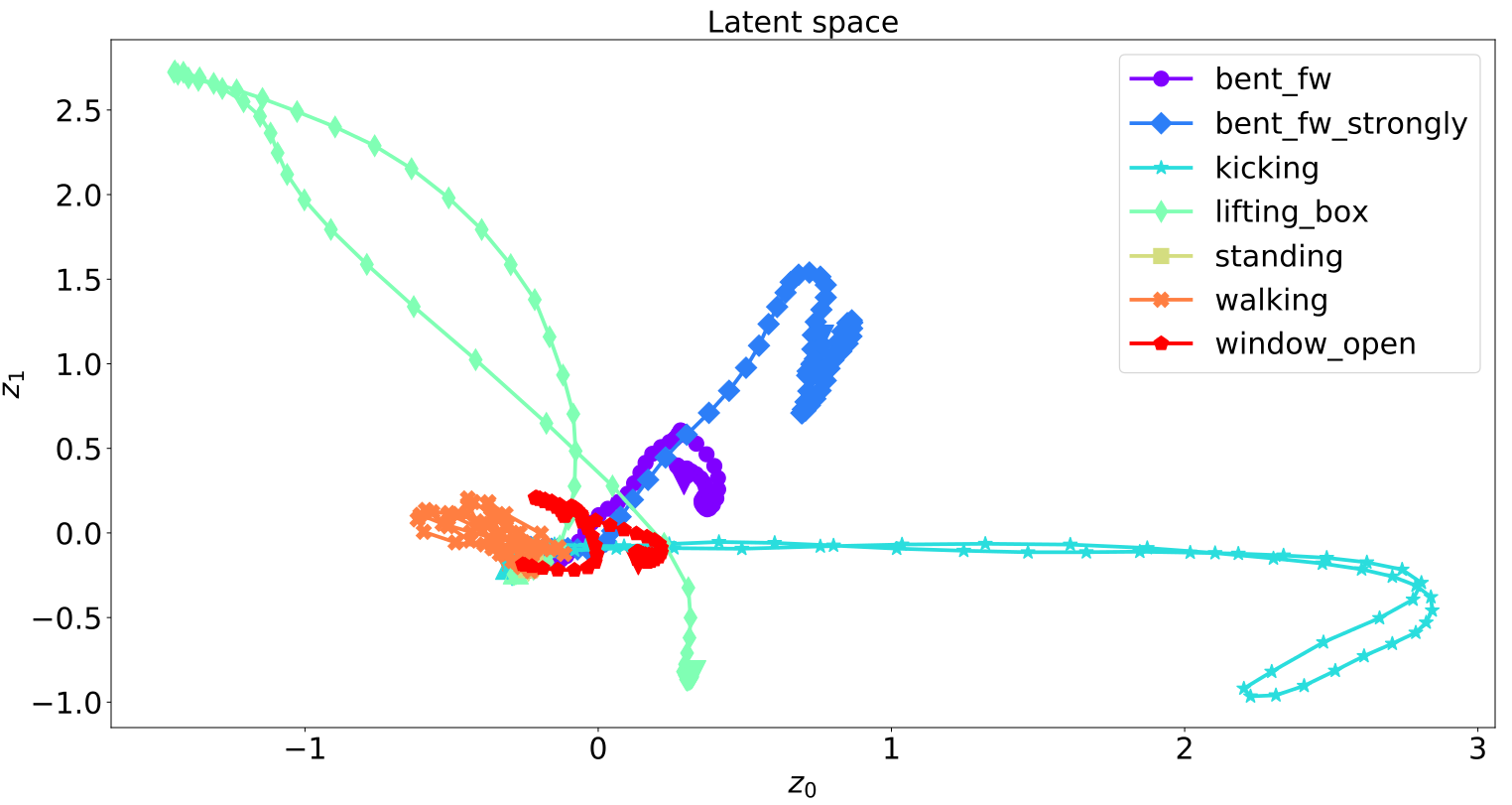}

(c)  \includegraphics[scale=0.16]{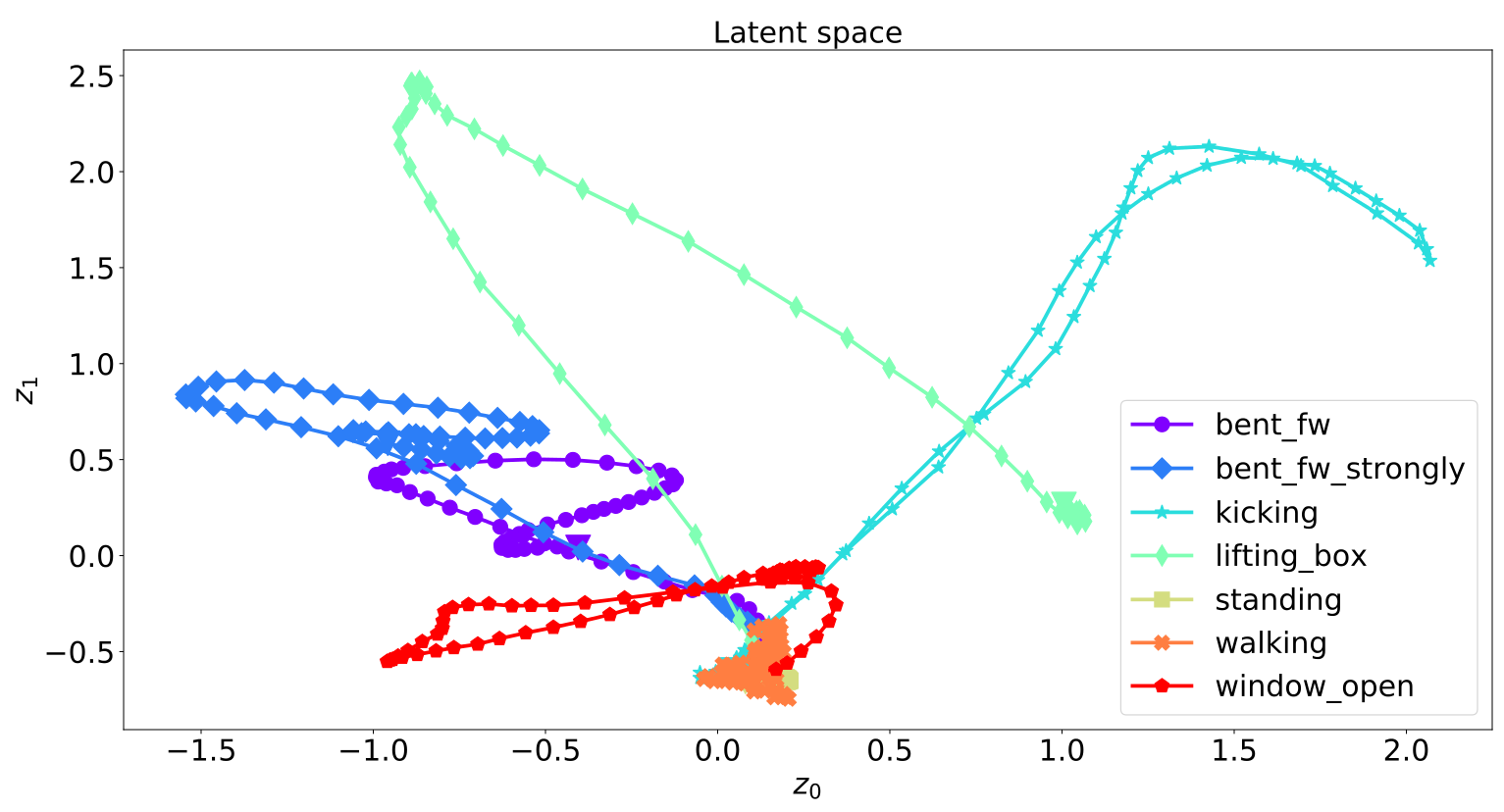}
  
  }
  \caption{Latent space of (a) VAE only \rom{4} (b) VAE-DMP \rom{3} (c) VTSFE light \rom{2} in 2D showing trajectories for a new movement sample for each movement type.}
  \label{fig:ls_all}
\end{figure}

Finally, the latent space of these models can provide complementary information about encoded dynamics. While the ($\sum var$) indicator accounts for dynamics amplitude, the latent space smoothness accounts for acceleration continuity, which is crucial for prediction. 
In VAE only \rom{4} (Figure \ref{fig:ls_all}a), trajectories make rough changes in speed and direction. 
In VAE-DMP \rom{3} (Figure \ref{fig:ls_all}b), trajectories are much smoother than VAE only \rom{4}, and better suited for prediction. However, as already observed in the input data space, it lacks information about dynamics amplitude.
In VTSFE light \rom{2} (Figure \ref{fig:ls_all}c), trajectories are smooth, compatible with prediction and wider than VAE-DMP \rom{3}. In particular, it can be observed much more easily in VTSFE light \rom{2} that ``bent forward'' (purple) and ``bent forward strongly'' (blue) are similar movement, the second one being an emphasized version of the first one. 
Table \ref{fig:computation_times} reports on computation times for 10 training epochs for the 4 models considered in Section \ref{sec:experiments} in the same conditions as in Table \ref{fig:results}, on a 64-bit laptop with 15.5 GB RAM, Intel Core i7-4900MQ CPU @ 2.80 GHz x 8 and NVIDIA Quadro K2100M.

\section{Discussion}\label{sec:discussion}
However, it could be argued that $\sum var$ isn't a robust indicator and can only suggest the encoded dynamics quality. In fact, while this suggestion matches the one found in latent space about encoded features, these results do not account for a better reconstruction capability in VTSFE than in VAE-DMP or VAE only. Our results strongly suggest a better feature encoding in latent space with VTSFE, i.e. a better encoder, than VAE-DMP or VAE only, but not a better reconstruction, i.e. a better decoder. 
Thus, it is very likely that a second training phase for VTSFE, in which the encoder weights would be frozen and the loss function would be the one of the classic VAE, would result in a better reconstruction capability as we would get rid of all encoding constraint in loss function and we would feed the decoder with stable input data, i.e. a constant latent space. But above all, this second training phase would probably be much more efficient if we remove completely this variational aspect, simply taking the loss function of a classic Auto-Encoder, as we would train the system to reconstruct one input value from one latent value, instead of trying to match several samples of latent space to this one input value. Nevertheless, that is not the scope of this paper which focuses on feature extraction (encoding).

Therefore, our results show that VTSFE \rom{2} seems more suitable than VAE-DMP \rom{3} for prediction and classification of actions based on its inferred latent space from motion capture/wearable sensing data. It is computationally faster, and it encodes a better dynamics of the trajectories, which leads to better reconstruction of high-amplitude joint motions (\textit{e.g.}, as in kicking). The latent space seems more consistent, trajectories in the latent space are smoother, and similar movements are closer (\textit{e.g.}, as in "bent" and "strongly bent").
\footnote{Our results could be biased by our implementation of VAE-DMP \cite{CheKarSma2016}, which makes some assumptions about the implementation that were not described in the original paper. }

\begin{table}[t]
	\centering
	\begin{tabular}{|l|p{1.cm}|c|}
	  \hline
	  & \emph{all} & \emph{separated} \\
	  \hline
	  VTSFE 2D $[P=15, L=15]$ & - & killed ($> 30$ GB RAM) \\
	  VTSFE 2D $[P=5, L=30]$ & - & 2h30m \\
	 VTSFE 2D $[P=5, L=15]$ & - & 1h30m \\
	  \hline
	  VTSFE light 2D $[L=30]$ & 9m & 9m30s \\
	  \hline
	  VAE-DMP 2D $[L=30]$ & 12m50 s & 16m30s \\
	  \hline
	  VAE only 2D, 5D, 7D $[L=30]$ & 5m40s & - \\
	  \hline
	\end{tabular}
	\caption{Computation times for 10 training epochs for the different learning schemes.}
	\vspace{-0.8cm}
	\label{fig:computation_times}
\end{table}

\section{Conclusion}\label{conclusion}

In this paper, we presented VTSFE (\acrlong{vtsfe}), that can be used to extract relevant features from motion capture/wearable sensing data to recognize and predict the current human activity. We compared it with our open-source implementation of VAE-DMP from \cite{CheKarSma2016} and classical VAE, showing that our algorithm is capable of better encoding the dynamics of the captured movements: this is an essential feature for our ongoing work, where VTSFE is needed for predicting the future trajectory of the human motion. 
Due to Monte Carlo samplings, the computation time for training is still non-negligible, even for our relatively small dataset, which makes it suitable only as a pre-computed model trained offline.
Nevertheless, once trained VTSFE is able to encode features online.

In the future, we plan several improvements of our method.
In particular, our model is agnostic of the type of input data. While this is a potential advantage (as we could input different types of sensor data), it does not exploit any prior knowledge about the human-body model.
We plan to improve the VAE part of our model by testing {Ladder Variational Autoencoders} \cite{lvae}, which outperform standard \acrshort{vae} in posterior approximation and are used to train a deep architecture of latent layers, and {VAE-GAN} \cite{vae-gan}, which outperform \acrshort{vae} by combining it with a Generative Adversarial Network (GAN) that enhances the plausibility of the generated data.

Also, as discussed before, we plan to focus on a second phase of training for the decoder part of VTSFE to enhance its reconstruction capability.

Finally, we would like to make a more thorough\footnote{Our evaluation so far has been significantly limited by the availability of a cluster for computations.} evaluation with a bigger dataset and a varying number of sensors, combining also different input sensors (e.g., IMU, EMG, etc.).

\addtolength{\textheight}{-2.5cm}   

\bibliographystyle{IEEEtran}
\bibliography{vtsfe}  

\onecolumn

\begin{appendices}

\section{Related works}

\subsection{VAE}

\begin{equation}
\begin{aligned}
\ln(p_{\theta}([x_k]_{k=1}^K)) &= \ln( \prod_{k=1}^{K} p_{\theta}(x_k))\\
&= \sum_{k=1}^{K} ln(p_{\theta}(x_k))
\end{aligned}
\end{equation}
\begin{equation}\label{vae_max}
\begin{aligned}
\ln(p_{\theta}(x_k)) &= \int_{z} q_{\phi}(z|x_k) \ln(p_{\theta}(x_k)) dz \\
&= \int_{z} q_{\phi}(z|x_k) \ln(\frac{p_{\theta}(z,x_k)}{q_{\phi}(z|x_k)}) + q_{\phi}(z|x_k) \ln(\frac{q_{\phi}(z|x_k)}{p_{\theta}(z|x_k)})~ dz\\
& = ELBO + D_{KL}(q_{\phi}(z|x_k)~||~p_{\theta}(z|x_k))
\end{aligned}
\end{equation}
where $ELBO$ is the Evidence Lower BOund.

\begin{equation}\label{vae_elbo}
\begin{aligned}
-ELBO &= -\int_{z} q_{\phi}(z|x_k) \ln(\frac{p_{\theta}(z,x_k)}{q_{\phi}(z|x_k)}) ~ dz \\
&= - \int_{z} q_{\phi}(z|x_k) \ln(p_{\theta}(x_k|z)) + q_{\phi}(z|x_k) \ln(\frac{p_{\theta}(z)}{q_{\phi}(z|x_k)}) ~ dz\\
& = - \mathbb{E}_{q_{\phi}(z|x_k)}[\ln(p_{\theta}(x_k|z))] + D_{KL}(q_{\phi}(z|x_k)~||~p_{\theta}(z))
\end{aligned}
\end{equation}

\subsection{DVBF}

\begin{equation}\label{dvbf_marg}
\begin{aligned}
p_{\theta}(x_{1:T} | u_{1:T}, z_1) &= \int_{z_{2:T}} p_{\theta}(x_{1:T} | z_{1:T}, u_{1:T}) p_{\theta}(z_{2:T} | u_{1:T}) dz_{2:T} \\
&= \int_{z_{2:T}} \int_{\beta_{1:T}} p_{\theta}(x_{1:T} | z_{1:T}, u_{1:T}) p_{\theta}(z_{2:T} | \beta_{1:T}, u_{1:T}) p_{\theta}(\beta_{1:T}) d\beta_{1:T} dz_{2:T}\\
&= \int_{z_{2:T}} \int_{\beta_{1:T}} p_{\theta}(\beta_{1:T}) \displaystyle \prod_{t=1}^{T} p_{\theta}(x_t | z_t) \displaystyle \prod_{t=2}^{T} p_{\theta}(z_{t} | z_{t-1}, u_{t-1}, \beta_{t-1}) d\beta_{1:T} dz_{2:T}
\end{aligned}
\end{equation}
Given the deterministic transition (\ref{eq:bayes_filt}), they noticed that $\displaystyle \prod_{t=2}^{T} p_{\theta}(z_{t} | z_{t-1}, u_{t-1}, \beta_{t-1})$ was a product of Dirac distributions and thus:
\begin{equation}
p_{\theta}(x_{1:T} | u_{1:T}, z_1) = \int_{\beta_{1:T}} p_{\theta}(\beta_{1:T}) p_{\theta}(x_t | z_1) \displaystyle \prod_{t=2}^{T} p_{\theta}(x_t | \hat{z}_t) ~d\beta_{1:T}
\end{equation}
where $\hat{z}_t=g(z_{t-1}, u_{t-1}, \beta_{t-1})$.

\begin{equation}
\begin{aligned}
& \ln(p_{\theta}(x_{1:T} | u_{1:T}, z_1)) \\
= & ~ \ln\left( \int_{\beta_{1:T}} \frac{q_{\phi}(\beta_{1:T} | x_{1:T}, u_{1:T}, z_1, \hat{z}_{2:T})}{q_{\phi}(\beta_{1:T} | x_{1:T}, u_{1:T}, z_1, \hat{z}_{2:T})} ~p_{\theta}(\beta_{1:T}) p_{\theta}(x_t | z_1) \displaystyle \prod_{t=2}^{T} p_{\theta}(x_t | \hat{z}_t) ~d\beta_{1:T} \right)\\
\geq & ~ \int_{\beta_{1:T}} q_{\phi}(\beta_{1:T} | x_{1:T}, u_{1:T}, z_1, \hat{z}_{2:T}) \ln \left(\frac{p_{\theta}(\beta_{1:T})}{q_{\phi}(\beta_{1:T} | x_{1:T}, u_{1:T}, z_1, \hat{z}_{2:T})} p_{\theta}(x_t | z_1) \displaystyle \prod_{t=2}^{T} p_{\theta}(x_t | \hat{z}_t)\right) ~d\beta_{1:T} = LB\\
\end{aligned}
\end{equation}
where:
\begin{equation}\label{eq:dvbf_lb}
\begin{aligned}
& LB = \mathbb{E}_{q_{\phi}(\beta_{1:T}|x_{1:T}, u_{1:T}, z_1, \hat{z}_{2:T})}\left[\ln\left(p_{\theta}(x_t | z_1) \displaystyle \prod_{t=2}^{T} p_{\theta}(x_t|\hat{z}_t)\right)\right] - D_{KL}(q_{\phi}(\beta_{1:T}|x_{1:T}, u_{1:T}, z_1, \hat{z}_{2:T})~\|~p_{\theta}(\beta_{1:T}))
\end{aligned}
\end{equation}

\section{Methods}
\label{sec:methods}

\subsection{Our improvements to VAE-DMP}\label{sec:vae-dmp-cor}

\subsubsection{Tighter lower bound}

\begin{equation}\label{eq:q_decomp_1}
\begin{aligned}
q_{\phi}(z_{1:T}|x_{1:T}) &= \frac{q_{\phi}(z_{1:T}, x_{1:T})}{q_{\phi}(x_{1:T})}\\
&= \displaystyle \prod_{t\in \lbrace{1,2,T}\rbrace} q_{\phi}(z_{t}|x_{t}) \displaystyle \prod_{t=3}^{T-1} q_{\phi}(z_t|x_{1:T}, z_{t-1}, z_{t-2}, z_{T})
\end{aligned}
\end{equation}
Given $\forall t \in [3,T-1], z_t = g(z_{t-1}, z_{t-2}, z_{T}, \epsilon_{t-1}, f_{t-1})$, where $g$ is a deterministic transition function, we can push (\ref{eq:q_decomp_1}) a little further :
\begin{equation}\label{eq:z_encoding}
\begin{aligned}
& q_{\phi}(z_{1:T}|x_{1:T})\\
&= \displaystyle \prod_{t\in \lbrace{1,2,T}\rbrace} q_{\phi}(z_{t}|x_{t}) \displaystyle \prod_{t=2}^{T-2} q_{\phi}(g(z_{t}, z_{t-1}, z_{T}, \epsilon_{t}, f_{t})|x_{1:T}, z_{t}, z_{t-1}, z_{T})\\
&= \displaystyle \prod_{t\in \lbrace{1,2,T}\rbrace} q_{\phi}(z_{t}|x_{t}) \displaystyle \prod_{t=2}^{T-2} q_{\phi}(\epsilon_{t}, f_{t}|x_{1:T}, z_{t}, z_{t-1}, z_{T})\\
&= \displaystyle \prod_{t\in \lbrace{1,2,T}\rbrace} q_{\phi}(z_{t}|x_{t}) \displaystyle \prod_{t=2}^{T-2} \frac{q_{\phi}(\epsilon_{t}, f_{t}, x_{1:T}, z_{t}, z_{t-1}, z_{T})}{q_{\phi}(x_{1:T}, z_{t}, z_{t-1}, z_{T})}\\
&= \displaystyle \prod_{t\in \lbrace{1,2,T}\rbrace} q_{\phi}(z_{t}|x_{t})  \displaystyle \prod_{t=2}^{T-2} \frac{q_{\phi}(f_{t}|x_{1:T}) q_{\phi}(\epsilon_{t} | f_t, x_{t+1}, z_{t}, z_{t-1}, z_{T}) q_{\phi}(x_{1:T}, z_{t}, z_{t-1}, z_{T})}{q_{\phi}(x_{1:T}, z_{t}, z_{t-1}, z_{T})}\\
&= \displaystyle \prod_{t\in \lbrace{1,2,T}\rbrace} q_{\phi}(z_{t}|x_{t})  \displaystyle \prod_{t=2}^{T-2}{q_{\phi}(\epsilon_{t}|f_t, x_{t+1}, z_{t}, z_{t-1}, z_{T}) q_{\phi}(f_{t}|x_{1:T})}\\
\end{aligned}
\end{equation}

In the same way:
\begin{equation}\label{eq:z_prior}
\begin{aligned}
p_{\theta}(z_{1:T}) &= \displaystyle \prod_{t\in \lbrace{1,2,T}\rbrace} p_{\theta}(z_t)  \displaystyle \prod_{t=3}^{T-1} p_{\theta}(z_t | z_{t-1}, z_{t-2}, z_{T})\\
&= \displaystyle \prod_{t\in \lbrace{1,2,T}\rbrace} p_{\theta}(z_t)  \displaystyle \prod_{t=2}^{T-2} p_{\theta}(\epsilon_{t}, f_{t} | z_{t}, z_{t-1}, z_{T})\\
&= \displaystyle \prod_{t\in \lbrace{1,2,T}\rbrace} p_{\theta}(z_t)  \displaystyle \prod_{t=2}^{T-2} p_{\theta}(\epsilon_{t}) p_{\theta}(f_{t})
\end{aligned}
\end{equation}

And finally:
\begin{equation}\label{eq:reco_x}
p_{\theta}(x_{1:T} | z_{1:T}) = \displaystyle \prod_{t=1}^{T} p_{\theta}(x_{t} | z_{t})
\end{equation}
(\ref{eq:q_decomp_1}) and (\ref{eq:reco_x}) lead to:
\begin{equation}\label{eq:new_lb_reco_err}
\begin{aligned}
&\mathbb{E}_{q_{\phi}(z_{1:T}|x_{1:T})}[\ln(p_{\theta}(x_{1:T}|z_{1:T}))]\\
&= \mathbb{E}_{q_{\phi}(z_{1:T}|x_{1:T})}\left[ \sum_{t=1}^{T} \ln(p_{\theta}(x_{t}|z_{t}))\right]\\
&= \sum_{t_1=1}^{T} \int_{z_{1:T}} \displaystyle \prod_{t_2\in \lbrace{1,2,T}\rbrace} q_{\phi}(z_{t_2}|x_{t_2}) \displaystyle \prod_{t_2=3}^{T-1} q_{\phi}(z_{t_2}|x_{1:T}, z_{t_2-1}, z_{t_2-2}, z_{T}) \ln(p_{\theta}(x_{t_1}|z_{t_1}))dz_{1:T} \\
&= \sum_{t_1\in \lbrace{1,2,T}\rbrace} \left[\int_{z_{t_1}} q_{\phi}(z_{t_1}|x_{t_1}) \ln(p_{\theta}(x_{t_1}|z_{t_1})) \right.\\
&\qquad \times \left [ \int_{z_{[1:T]\backslash\lbrace{t_1}\rbrace}} \displaystyle \prod_{t_2\in \lbrace{1,2,T}\rbrace\backslash\lbrace{t_1}\rbrace} q_{\phi}(z_{t_2}|x_{t_2}) \right.\\
&\left.\left. \qquad\quad\times \displaystyle \prod_{t_2=3}^{T-1} q_{\phi}(z_{t_2}|x_{1:T}, z_{t_2-1}, z_{t_2-2}, z_{T}) ~dz_{[1:T]\backslash\lbrace{t_1}\rbrace} \right] ~dz_{t_1} \right]\\
& \quad + \sum_{t_1=3}^{T-1} \left [\int_{z_{1:t_1},z_T} \displaystyle \prod_{t_2\in \lbrace{1,2,T}\rbrace} q_{\phi}(z_{t_2}|x_{t_2}) \displaystyle \prod_{t_2=3}^{t_1} q_{\phi}(z_{t_2}|x_{1:T}, z_{t_2-1}, z_{t_2-2}, z_{T}) \ln(p_{\theta}(x_{t_1}|z_{t_1})) \right.\\
& \left.\qquad \times \left [ \int_{z_{[t_1+1:T-1]}}  \displaystyle \prod_{t_2=t_1+1}^{T-1} q_{\phi}(z_{t_2}|x_{1:T}, z_{t_2-1}, z_{t_2-2}, z_{T}) ~dz_{[t_1+1:T-1]} \right] ~dz_{1:t_1} dz_T \right]\\
&= \sum_{t\in \lbrace{1,2,T}\rbrace} \left[\int_{z_{t}} q_{\phi}(z_{t}|x_{t}) \ln(p_{\theta}(x_{t}|z_{t})) ~dz_{t} \right]\\
& \quad + \sum_{t=3}^{T-1} \left [\int_{z_{1:t},z_T} q_{\phi}(z_{1:t},z_T|x_{1:T}) \ln(p_{\theta}(x_{t}|z_{t})) ~dz_{1:t} dz_T \right]\\
&= \sum_{t\in \lbrace{1,2,T}\rbrace} \mathbb{E}_{q_{\phi}(z_{t}|x_{t})}[ \ln(p_{\theta}(x_{t}|z_{t}))] + \sum_{t=3}^{T-1} \mathbb{E}_{q_{\phi}(z_{1:t},z_T|x_{1:T})}[ \ln(p_{\theta}(x_{t}|z_{t}))]\\
\end{aligned}
\end{equation}
Besides, (\ref{eq:z_prior}) and (\ref{eq:z_encoding}) lead to:
\begin{equation}\label{eq:dkl_decomp}
\begin{aligned}
& D_{KL}(q_{\phi}(z_{1:T}|x_{1:T})~||~p_{\theta}(z_{1:T})) \\
&= \int_{z_{1:T}} q_{\phi}(z_{1:T}|x_{1:T}) \ln\left(\frac{q_{\phi}(z_{1:T}|x_{1:T})}{p_{\theta}(z_{1:T})}\right) ~ dz_{1:T}\\
&= \int_{z_{1:T}} q_{\phi}(z_{1:T}|x_{1:T}) \ln \left(\frac{ \displaystyle \prod_{t\in \lbrace{1,2,T}\rbrace} q_{\phi}(z_{t}|x_{t}) \displaystyle \prod_{t=2}^{T-2} q_{\phi}(\epsilon_{t}|f_t, x_{t+1}, z_{t}, z_{t-1}, z_{T}) q_{\phi}(f_{t}|x_{1:T})}{\displaystyle \prod_{t\in \lbrace{1,2,T}\rbrace} p_{\theta}(z_t) \displaystyle \prod_{t=2}^{T-2} p_{\theta}(\epsilon_{t}) p_{\theta}(f_{t})} \right) ~ dz_{1:T}\\
&= \int_{z_{1:T}} q_{\phi}(z_{1:T}|x_{1:T}) \left[ \displaystyle \sum_{t\in \lbrace{1,2,T}\rbrace} \ln \left (\frac{q_{\phi}(z_{t}|x_{t})}{p_{\theta}(z_t)}\right) + \sum_{t=2}^{T-2} \ln\left(\frac{q_{\phi}(\epsilon_{t}|f_t, x_{t+1}, z_{t}, z_{t-1}, z_{T})}{p_{\theta}(\epsilon_{t})}\right) \right.\\
&\left. \quad \quad + \sum_{t=2}^{T-2} \ln(q_{\phi}(f_{t}|x_{1:T})) - \sum_{t=2}^{T-2} \ln(p_{\theta}(f_{t}))\right] ~ dz_{1:T}\\
&= \sum_{t\in \lbrace{1,2,T}\rbrace} \int_{z_{t}} q_{\phi}(z_{t}|x_{t}) \ln \left(\frac{q_{\phi}(z_{t}|x_{t})}{p_{\theta}(z_t)}\right) dz_t \\
&\quad + \sum_{t_1=2}^{T-2} \left[ \int_{z_{1:t_1}, z_T} \int_{\epsilon_{t_1}} \int_{f_{t_1}} \displaystyle \prod_{t_2\in \lbrace{1,2,T}\rbrace} q_{\phi}(z_{t_2}|x_{t_2}) \displaystyle \prod_{t_2=3}^{t_1} q_{\phi}(z_{t_2}|x_{1:T}, z_{t_2-1}, z_{t_2-2}, z_{goal}) \right. \\
& \qquad \times q_{\phi}(\epsilon_{t_1}|f_{t_1}, x_{t_1+1}, z_{t_1}, z_{t_1-1}, z_{T}) q_{\phi}(f_{t_1} | x_{1:T})~\left[ \ln\left(\frac{q_{\phi}(\epsilon_{t_1}|f_{t_1}, x_{t_1+1}, z_{t_1}, z_{t_1-1}, z_{T})}{p_{\theta}(\epsilon_{t_1})}\right) \right.\\
& \left. \qquad \qquad + \ln(q_{\phi}(f_{t_1} | x_{1:T})) - \ln(p_{\theta}(f_{t_1})) \vphantom{\frac{q_{\phi}(\epsilon_{t_1}|f_{t_1}, x_{t_1+1}, z_{t_1}, z_{t_1-1}, z_{T})}{p_{\theta}(\epsilon_{t_1})}}\right]\\
& \left. \qquad \times \left[ \int_{z_{[t_1+1:T-1]}} \displaystyle \prod_{t_2=t_1+1}^{T-1} q_{\phi}(z_{t_2}|x_{1:T}, z_{t_2-1}, z_{t_2-2}, z_{goal}) ~dz_{[t_1+1:T-1]} \right] ~df_{t_1}d\epsilon_{t_1} dz_{1:t_1} dz_T \right]\\
&= \sum_{t\in \lbrace{1,2,T}\rbrace} D_{KL}(q_{\phi}(z_{t}|x_{t})~||~p_{\theta}(z_{t})) \\
&\quad + \sum_{t_1=2}^{T-2} \mathbb{E}_{q_{\phi}(z_{1:t_1}, z_T|x_{1:T})} \left[ \int_{f_{t_1}} q_{\phi}(f_{t_1} | x_{1:T}) \right.\\
& \qquad \left[ \int_{\epsilon_{t_1}} q_{\phi}(\epsilon_{t_1}|f_{t_1}, x_{t_1+1}, z_{t_1}, z_{t_1-1}, z_{T})~ \ln\left(\frac{q_{\phi}(\epsilon_{t_1}|f_{t_1}, x_{t_1+1}, z_{t_1}, z_{t_1-1}, z_{T})}{p_{\theta}(\epsilon_{t_1})}\right)  ~ d\epsilon_{t_1} \right.\\
& \qquad + \ln(q_{\phi}(f_{t_1} | x_{1:T})) \int_{\epsilon_{t_1}} q_{\phi}(\epsilon_{t_1}|f_{t_1}, x_{t_1+1}, z_{t_1}, z_{t_1-1}, z_{T}) ~d\epsilon_{t_1}\\
& \left.\left. \qquad - \ln(p_{\theta}(f_{t_1})) \int_{\epsilon_{t_1}} q_{\phi}(\epsilon_{t_1}|f_{t_1}, x_{t_1+1}, z_{t_1}, z_{t_1-1}, z_{T}) d\epsilon_{t_1} \right] df_{t_1} \right]\\
\end{aligned}
\end{equation}

As $f_{t}$ is inferred deterministically by $x_{1:T}$, the probability of $f_{t}$ given $x_{1:T}$ equals $1$. Thus, the distribution $q_{\phi}(f_{t}|x_{1:T})$ is a Dirac function centered at $\hat{f}_t$, where $\hat{f}_t$ is the inferred value of $f_t$ for a given $x_{1:T}$. Therefore, $\int_{f_{t_1}} q_{\phi}(f_{t_1} | x_{1:T})~ \ln(q_{\phi}(f_{t_1} | x_{1:T})) ~df_{t_1} = \ln(1) = 0$ and $\int_{f_{t_1}} q_{\phi}(f_{t_1} | x_{1:T})~ \ln(p_{\theta}(f_{t_1})) ~df_{t_1} = \ln(p_{\theta}(\hat{f}_t))$. Thereby, (\ref{eq:dkl_decomp}) simplifies as following:
\begin{equation}
\begin{aligned}
& D_{KL}(q_{\phi}(z_{1:T}|x_{1:T})~||~p_{\theta}(z_{1:T})) \\
&= \sum_{t\in \lbrace{1,2,T}\rbrace} D_{KL}(q_{\phi}(z_{t}|x_{t})~||~p_{\theta}(z_{t})) \\
&\quad + \sum_{t_1=2}^{T-2} \mathbb{E}_{q_{\phi}(z_{1:t_1}, z_T|x_{1:T})} \left[ - \ln(p_{\theta}(\hat{f}_{t_1})) \vphantom{\frac{q_{\phi}(\epsilon_{t_1}|\hat{f}_{t_1}, x_{t_1+1}, z_{t_1}, z_{t_1-1}, z_{T})}{p_{\theta}(\epsilon_{t_1})}}\right.\\
& \left. \qquad + \int_{\epsilon_{t_1}} q_{\phi}(\epsilon_{t_1}|\hat{f}_{t_1}, x_{t_1+1}, z_{t_1}, z_{t_1-1}, z_{T})~ \ln\left(\frac{q_{\phi}(\epsilon_{t_1}|\hat{f}_{t_1}, x_{t_1+1}, z_{t_1}, z_{t_1-1}, z_{T})}{p_{\theta}(\epsilon_{t_1})}\right)  ~ d\epsilon_{t_1} \right]\\
&= \sum_{t\in \lbrace{1,2,T}\rbrace} D_{KL}(q_{\phi}(z_{t}|x_{t})~||~p_{\theta}(z_{t})) \\
&\quad + \sum_{t=2}^{T-2} \mathbb{E}_{q_{\phi}(z_{1:t}, z_T|x_{1:T})} \left[ - \ln(p_{\theta}(\hat{f}_{t})) \right.\\
& \left. \qquad + D_{KL}(q_{\phi}(\epsilon_{t}|\hat{f}_{t}, x_{t+1}, z_{t}, z_{t-1}, z_{T}) ~||~ p_{\theta}(\epsilon_{t})) \right]\\
\end{aligned}
\end{equation}

Light version calculation of $\displaystyle D_{KL}(q_{\phi}(\epsilon_{t}|f_{t}, x_{t+1}, z_{t}, z_{t-1}, z_{T}) ~||~ p_{\theta}(\epsilon_{t}))$ :
\begin{equation}\label{eq:dkl_noise}
\begin{aligned}
& D_{KL}(q_{\phi}(\epsilon_{t}|f_{t}, x_{t+1}, z_{t}, z_{t-1}, z_{T}) ~||~ p_{\theta}(\epsilon_{t})) \\
&= \frac{1}{2}\left[\ln\left(\frac{\left|\sigma_{scale}^2I\right|}{\left|\sigma_{scale}^2 I \sigma_{\epsilon,t}^2I\right|}\right) - d_z + \tr\left((\sigma_{scale}^2I)^{-1} \sigma_{scale}^2 I \sigma_{\epsilon,t}^2I\right) + (- \sigma_{scale} \odot \mu_{\epsilon,t}) (\sigma_{scale}^2 I)^{-1} (- \sigma_{scale} \odot \mu_{\epsilon,t})^T \right]\\
&= \frac{1}{2} \sum_{i=1}^{d_z} \left[ - \ln(\sigma_{\epsilon,t,i}^2) - 1 + \sigma_{\epsilon,t,i}^2 + \mu_{\epsilon,t,i}^2\right]\\
\end{aligned}
\end{equation}

The final $ELBO$ is thus defined as following :
\begin{equation}
\begin{aligned}
ELBO &=  \mathbb{E}_{q_{\phi}(z_{1:T}|x_{1:T})}[\ln(p_{\theta}(x_{1:T}|z_{1:T}))] - D_{KL}(q_{\phi}(z_{1:T}|x_{1:T})~||~p_{\theta}(z_{1:T}))\\
&\leq \bigg(\sum_{t\in \lbrace{1,2,T}\rbrace} \mathbb{E}_{q_{\phi}(z_{t}|x_{t})}[ \ln(p_{\theta}(x_{t}|z_{t}))]\\
& \qquad+ \sum_{t=3}^{T-1} \bigg[ \mathbb{E}_{q_{\phi}(z_{1:t},z_T|x_{1:T})}[ \ln(p_{\theta}(x_{t}|z_{t}))] \\
& \qquad+ \mathbb{E}_{q_{\phi}(z_{1:T}|x_{1:T})}[\ln(p_{\theta}(\hat{f}_{t-1} | z_{1:T}))] \bigg] \bigg)\\
&\quad - \bigg (\sum_{t\in \lbrace{1,2,T}\rbrace} D_{KL}(q_{\phi}(z_{t}|x_{t})~||~p_{\theta}(z_{t})) \\
&\qquad + \sum_{t=2}^{T-2} \mathbb{E}_{q_{\phi}(z_{1:t}, z_T|x_{1:T})} \bigg [D_{KL}(q_{\phi}(\epsilon_{t}|\hat{f}_{t}, x_{t+1}, z_{t}, z_{t-1}, z_{T}) ~||~ p_{\theta}(\epsilon_{t})) \bigg] \bigg)\\
&\leq \bigg(\sum_{t\in \lbrace{1,2,T}\rbrace} \mathbb{E}_{q_{\phi}(z_{t}|x_{t})}[ \ln(p_{\theta}(x_{t}|z_{t}))]\\
& \qquad+ \sum_{t=3}^{T-1} \mathbb{E}_{q_{\phi}(z_{1:t},z_T|x_{1:T})}[ \ln(p_{\theta}(x_{t}|z_{t}))] \\
&\quad - \bigg (\sum_{t\in \lbrace{1,2,T}\rbrace} D_{KL}(q_{\phi}(z_{t}|x_{t})~||~p_{\theta}(z_{t})) \\
&\qquad + \sum_{t=2}^{T-2} \mathbb{E}_{q_{\phi}(z_{1:t}, z_T|x_{1:T})} \bigg [D_{KL}(q_{\phi}(\epsilon_{t}|\hat{f}_{t}, x_{t+1}, z_{t}, z_{t-1}, z_{T}) ~||~ p_{\theta}(\epsilon_{t})) \bigg] \bigg)\\
&\approx \bigg(\sum_{t=1}^{T} \mathbb{E}_{q_{\phi}(z_t|x_t)}[\ln(p_{\theta}(x_{t}|z_{t}))]\bigg)\\
&\quad - \bigg ( \sum_{t\in \lbrace{1,2,T}\rbrace} D_{KL}(q_{\phi}(z_{t}|x_{t})~||~p_{\theta}(z_{t}))\\
&\qquad + \sum_{t=3}^{T-1} D_{KL}(q_{\phi}(\epsilon_{t}|\hat{f}_{t}, x_{t+1}, z_{t}, z_{t-1}, z_{T}) ~||~ p_{\theta}(\epsilon_{t})) \bigg)\\
\end{aligned}
\end{equation}
where the latter expression is equivalent to the lower bound of VAE-DMP.

\end{appendices}

\end{document}